\documentclass[10pt]{article} 

\usepackage[accepted]{tmlr}

\usepackage{algorithm}

\usepackage[utf8]{inputenc}
\usepackage{booktabs}
\usepackage{amsfonts}
\usepackage{nicefrac}
\usepackage{microtype}
\usepackage{xcolor}

\usepackage[noend]{algpseudocode}

\usepackage{amsmath}
\usepackage{amssymb}
\usepackage{siunitx}
\usepackage{amsthm}
\usepackage{fnpct}
\usepackage{hyperref}
\hypersetup{
    colorlinks,
    linkcolor={red},
    citecolor={blue},
    urlcolor={blue}
}
\usepackage{url}
\usepackage[capitalise,noabbrev]{cleveref}
\usepackage{mathtools}

\usepackage{xspace}
\usepackage{placeins}
\usepackage{tabularx}
\usepackage{pifont}
\usepackage[nolist]{acronym}
\usepackage{subcaption}

\usepackage{wrapfig}
\usepackage{placeins}

\makeatletter
\newcommand*{\org@overidelabel}{}
\let\org@overridelabel\@verridelabel
\@ifpackagelater{acronym}{2015/03/21}{%
  \renewcommand*{\@verridelabel}[1]{%
    \@bsphack
    \protected@write\@auxout{}{\string\AC@undonewlabel{#1@cref}}%
    \org@overridelabel{#1}%
    \@esphack
  }%
}{%
  \renewcommand*{\@verridelabel}[1]{%
    \@bsphack
    \protected@write\@auxout{}{\string\undonewlabel{#1@cref}}%
    \org@overridelabel{#1}%
    \@esphack
  }%
}
\makeatother

\newcommand{\polter}{\ac{POLTER}\xspace}

\DeclarePairedDelimiterX{\infdivx}[2]{(}{)}{%
  #1\;\delimsize\|\;#2%
}
\newcommand{\infdiv}{D_{\text{KL}}\infdivx}

\newcommand{\E}{\mathop{\mathbb{E}}}

\DeclareSIUnit{\nothing}{\relax}
\definecolor{blue}{HTML}{526fae}
\definecolor{green}{HTML}{63a76b}

\newcommand{\tick}{\ding{52}}

\newcommand{\cross}{\ding{56}}

\newcommand{\revt}[1]{\textcolor{black}{#1}}
\newenvironment{rev}{\color{black}}{\ignorespacesafterend}

\newcommand{\revteditor}[1]{\textcolor{black}{#1}}
\newenvironment{reveditor}{\color{black}}{\ignorespacesafterend}

\newenvironment{hproof}{%
  \proof}{\endproof}

\newtheorem{lemma}{Lemma}
\newtheorem{proposition}{Proposition}

\title{POLTER: Policy Trajectory Ensemble Regularization \\ for Unsupervised Reinforcement Learning}

\author{\name Frederik Schubert \email schubert@tnt.uni-hannover.de\\
  \addr Institute for Information Processing\\
  Leibniz University Hannover
  \AND
  \name Carolin Benjamins \email benjamins@ai.uni-hannover.de\\
  \addr Institute of Artificial Intelligence\\
  Leibniz University Hannover
  \AND
  \name Sebastian Döhler \email doehler@tnt.uni-hannover.de\\
  \addr Institute for Information Processing\\
  Leibniz University Hannover
  \AND
  \name Bodo Rosenhahn \email rosenhahn@tnt.uni-hannover.de\\
  \addr Institute for Information Processing\\
  Leibniz University Hannover
  \AND
  \name Marius Lindauer \email m.lindauer@ai.uni-hannover.de\\
  \addr Institute of Artificial Intelligence\\
  Leibniz University Hannover
}

\begin{acronym}
\acro{RL}{Reinforcement Learning}
\acro{MDP}{Markov Decision Process}
\acro{URL}{Unsupervised Reinforcement Learning}
\acro{URLB}{Unsupervised Reinforcement Learning Benchmark}
\acro{IQM}{Interquartile Mean}
\acro{POLTER}{Policy Trajectory Ensemble Regularization}
\acro{RND}{Random Network Distillation}
\acro{ICM}{Intrinsic Curiosity Module}
\acro{APT}{Active Pre-training}
\acro{APS}{Active Pre-training with Successor Features}
\acro{SMM}{State Marginal Matching}
\acro{CIC}{Contrastive Intrinsic Control}
\acro{DIAYN}{"Diversity is All You Need"}
\acro{DDPG}{Deep Deterministic Policy Gradient}
\acro{SAC}{Soft Actor-Critic}
\acro{ProtoRL}{ProtoRL}
\acro{KL-divergence}{Kullback–Leibler divergence}
\acro{Disagreement}{Disagreement}
\end{acronym}

\def\month{04}
\def\year{2023}
\def\openreview{\url{https://openreview.net/forum?id=Hnr23knZfY}}

\begin{document}

\maketitle

\begin{abstract}
    The goal of \ac{URL} is to find a reward-agnostic prior policy on a task domain, such that the sample-efficiency on supervised downstream tasks is improved.
    Although agents initialized with such a prior policy can achieve a significantly higher reward with fewer samples when finetuned on the downstream task, it is still an open question how an optimal pretrained prior policy can be achieved in practice.
    In this work, we present \acsu{POLTER} (\acl{POLTER}) – a general method to regularize the pretraining that can be applied to any \ac{URL} algorithm and is especially useful on data- and knowledge-based \ac{URL} algorithms.
    It utilizes an ensemble of policies that are discovered during pretraining and moves the policy of the \ac{URL} algorithm closer to its optimal prior.
    Our method is based on a theoretical framework, and we analyze its practical effects on a white-box benchmark, allowing us to study \polter with full control.
    In our main experiments, we evaluate \polter on the \ac{URLB}, which consists of 12 tasks in 3 domains.
    We demonstrate the generality of our approach by improving the performance of a diverse set of data- and knowledge-based \ac{URL} algorithms by \num{19}\% on average and up to \num{40}\% in the best case.
    Under a fair comparison with tuned baselines and tuned \ac{POLTER}, we establish a new state-of-the-art for model-free methods on the \ac{URLB}. 
\end{abstract}

\acresetall

\section{Introduction}

\ac{RL} has shown many successes in recent years \citep{mnihHumanlevelControlDeep2015,silverMasteringChessShogi2017,openaiSolvingRubikCube2019} and is starting to have an impact on real-world applications \citep{bellemareAutonomousNavigationStratospheric2020,mirhoseiniGraphPlacementMethodology2021,degraveMagneticControlTokamak2022}.
However, all these applications require detailed knowledge of the task to train the agents in a sufficiently close simulation.
Implementing each and every task and training a new agent is time-consuming and inefficient.
In some cases, the task might be difficult to model in simulation, or it is unknown beforehand, which makes sample-efficiency even more important.
\acf{URL}, i.e., reward-free pretraining in a task domain~\citep{srinivasUnsupervisedLearningReinforcement2021}, has shown to improve the sample-efficiency of finetuning \ac{RL} algorithms on downstream tasks.
This setup only requires a simulation of the agent's domain without having to model task-specific components in detail.
Although \ac{URL} results in a higher sample-efficiency on some applications, pretraining a policy still does not guarantee an improvement in performance on a wide range of tasks.
For example, \citet{laskinURLBUnsupervisedReinforcement2021} observed that longer pretraining often degrades the performance of many \ac{URL} algorithms.
Also, many finetuning steps are still needed to achieve the optimal return, and some algorithms fail to learn a sufficient policy at all. 
\begin{figure*}[t!]
    \centering
    \includegraphics[width=0.9\linewidth]{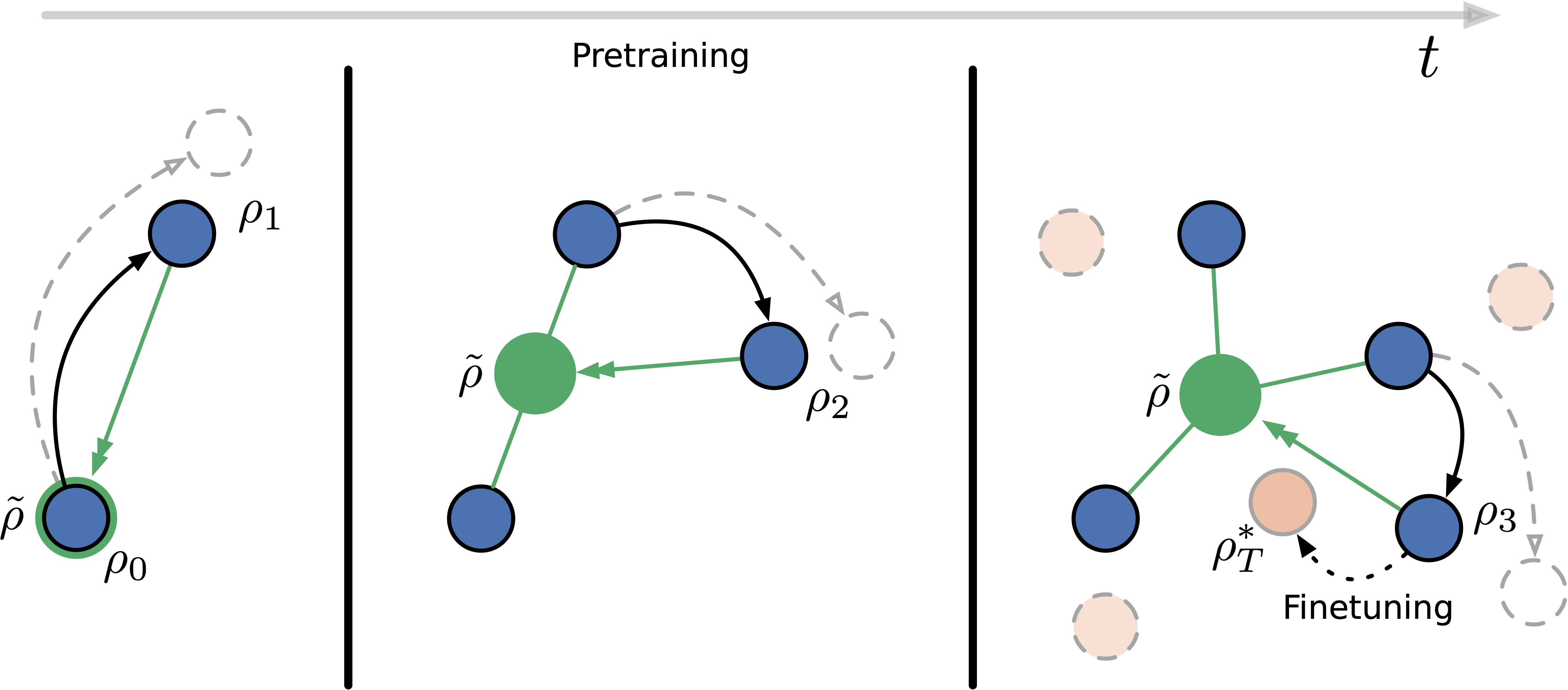}
    \caption{
    Optimization trajectory in the induced state-marginal space $\rho(s)$ of a policy $\pi(s)$ that optimizes an \aclu{URL} objective with \polter regularization.
    The regularized policy's state distribution $\rho_t(s)$ \includegraphics[height=8pt]{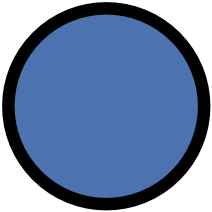} is pulled towards the ensemble state distribution $\tilde{\rho}(s)$ \includegraphics[height=8pt]{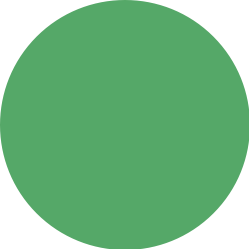} during pretraining.
    The ensemble policy $\tilde{\pi}(s)$ approximates the optimal prior $\pi^*_T(s)$ with its induced state-marginal $\rho^*_T(s)$ \includegraphics[height=8pt]{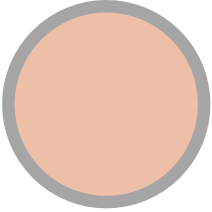} for finetuning on a specific task $T$.
    Note that other tasks \includegraphics[height=8pt]{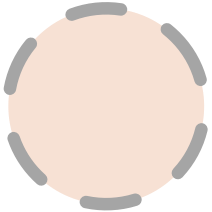} might lie further away from the average.
    However, as shown by \citet{eysenbachInformationGeometryUnsupervised2022}, the average still minimizes the distance without knowledge about the specific downstream task.
    The state distribution trajectory of a policy without \polter regularization is indicated by dashed lines \includegraphics[height=8pt]{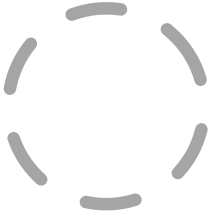}.
    }
    \label{fig:motivation}
\end{figure*}

In this work, we introduce \acsu{POLTER} (\acl{POLTER}) – a general method to improve the performance of \ac{URL} algorithms via a novel regularization term during pretraining that results in better prior policies for finetuning on specific tasks.
Our theoretical motivation builds on the geometric interpretation of \ac{URL} by~\citet{eysenbachInformationGeometryUnsupervised2022} who connect the optimal prior policy to a specific state distribution.
\begin{reveditor}
    The prior state distribution is optimal in the sense that it minimizes the distance to the state distribution of an adversarially chosen downstream task.
    This problem statement can be seen an instance of the information-theoretic framework of bounded rationality \citep{ortegaAdversarialInterpretationInformationTheoretic2014}.
\end{reveditor}
Some \ac{URL} algorithms try to approximate the optimal prior state distribution by learning separate skills during pretraining that are combined to initialize the finetuning policy.
\polter borrows this idea for \ac{URL} algorithms that do not learn explicit skills, e.g., data- and knowledge-based \ac{URL} algorithms.
This is based on the insight that each point during pretraining encodes an implicit skill that depends on the pretraining objective.
In a nutshell, our main idea is to minimize the distance of the current policy to an ensemble policy during pretraining, which acts as a proxy for the optimal prior.
As shown in \cref{fig:motivation}, the pretraining trajectory of the policy and hence the induced state distribution is affected by our regularization and attracted to the ensemble of earlier pretraining policies, each potentially encoding a different skill.

We demonstrate the effect of \polter on the simplistic PointMass environment and two tasks in the Pendulum domain, and provide empirical evidence that our method reduces the distance to the optimal prior policy and improves the downstream performance.
Extensive experiments on the \ac{URLB} \citep{laskinURLBUnsupervisedReinforcement2021} show that our method improves the \ac{IQM} \citep{agarwalDeepReinforcementLearning2021} performance of data- and knowledge-based \ac{URL} algorithms on average by 19\%.
A sensitivity analysis reveals that \polter is a robust method with a reasonable default setting but can further be tuned on some domains and algorithms.
Additional experiments on the state-visitation entropy provide empirical evidence that our method improves the performance by finding better priors in contrast to an improved exploration.
In summary, our \textbf{contributions} are:

\begin{enumerate}
    \item We introduce \polter, a novel regularization method for \ac{URL} algorithms that uses an ensemble of pretraining policies to reduce the deviation from the optimal prior policy.
    \item Our method is empirically validated on the white-box PointMass benchmark and demonstrated in the Pendulum domain.
    \item An extensive evaluation on the \ac{URLB} shows the effectiveness of \polter for a broad range of \ac{URL} algorithms with \revteditor{pixel- and state-based observations}.
    \item With our sensitivity analysis, we demonstrate that our approach is robust with a good default and can be further boosted via a task-specific strength of the regularization and achieves a new state-of-the-art for model-free algorithms on the \ac{URLB}.
\end{enumerate} 
\section{Preliminaries}

\subsection{Notation in Reinforcement Learning (RL)}

\ac{RL} relies on the notion of a \ac{MDP} \citep{suttonReinforcementLearningIntroduction1998}.
An \ac{MDP} is a tuple $(\mathcal{S}, \mathcal{A}, \mathcal{P}, r, \gamma)$ which describes the states $s \in \mathcal{S}$, actions $a \in \mathcal{A}$, transition dynamics $p(s_{t + 1} \mid s_t, a_t) \sim \mathcal{P}$, initial state distribution $p_0(s_0) \sim \mathcal{P}$, reward function $r(s_t)$ and discount factor $\gamma \in (0, 1]$ of an environment.
At each time step $t$, an agent observes the state $s_t$ of the environment and chooses an action $a_t$ using a policy $\pi(a_t \mid s_t)$ to transition into a new state $s_{t + 1}$.
The objective is to maximize the expected discounted return by finding a policy $\pi^*(s)$ that induces the optimal discounted state occupancy measure \revt{or state-marginal} $\rho^*(s) = (1 - \gamma) \sum_{t = 0}^{\infty} \gamma^t P^{\pi^*}_{t}(s)$, where $P^{\pi^*}_{t}(s)$ is the probability of being in state $s$ at time step $t$ when following policy $\pi^*$.
As we focus on state-based reward functions $r(s)$, this optimal discounted state occupancy measure corresponds to the maximal discounted return of the optimal policy.
 
\subsection{Related Work}

\paragraph{\acl{URL}} describes algorithms that train an \ac{RL} agent without a learning signal, i.e., without a task-specific reward function.
It encompasses two main research branches: The first is Unsupervised Representation Learning \citep{oordRepresentationLearningContrastive2019,ebertVisualForesightModelBased2018,haRecurrentWorldModels2018,laskinCURLContrastiveUnsupervised2020,schwarzerDataEfficientReinforcementLearning2021,stookeDecouplingRepresentationLearning2021} which focuses on learning a representation of the states or dynamics from noisy or incomplete observations.
\revt{This line of work includes algorithms that learn world models to train an agent with imagined trajectories to reduce the number of environment interactions \citep{dockhornForwardModelApproximation2018,hansenTemporalDifferenceLearning2022,xuLearningGeneralWorld2022,rajeswarUnsupervisedModelbasedPretraining2022,seoReinforcementLearningActionfree2022}.}
The second branch is called Unsupervised Behavioral Learning.

\paragraph{Unsupervised Behavioral Learning} is a variant of the \ac{RL} problem where the agent can interact with the environment in a reward-free setting before being assigned the downstream task~$T$ defined by an external reward function~\citep{oudeyerIntrinsicMotivationSystems2007,jinRewardFreeExplorationReinforcement2020}.
During pretraining, intrinsic rewards are used to train a policy to become an optimal prior and initialization for the subsequent finetuning task.
We denote the policy after finetuning as $\pi^*_T$ and the corresponding oracle policy \revt{that acts optimally} on task $T$ as $\pi^+_T$.
Algorithms for Unsupervised Behavioral Learning can be categorized into knowledge-, data- and competence-based approaches~\citep{srinivasUnsupervisedLearningReinforcement2021}.
\textbf{Knowledge-based} algorithms define a self-supervised task by making predictions on some aspect of the environment~\revt{\citep{pathakCuriositydrivenExplorationSelfsupervised2017,chenNuclearNormMaximization2022,chenUncertaintyEstimationBased2022}}.
The approach \textbf{data-based} methods follow is to maximize the state visitation entropy to explore the environment~\citep{zhangExplorationMaximizingEnyi2020,guoGeometricEntropicExploration2021,muttiTaskAgnosticExplorationPolicy2021,muttiUnsupervisedReinforcementLearning2022,hazanProvablyEfficientMaximum2019,zhaoMixtureSurprisesUnsupervised2022,jacqC3POLearningAchieve2022,nedergaardKMeansMaximumEntropy2022}.
\revt{A recent approach \citep{heWassersteinUnsupervisedReinforcement2022} maximizes the Wasserstein distance of the induced state-marginals instead.}
\textbf{Competence-based} algorithms learn skills that maximize the mutual information between the trajectories and a space of skills~\revt{\citep{mohamedVariationalInformationMaximisation2015,gregorVariationalIntrinsicControl2017,baumliRelativeVariationalIntrinsic2021,jiangUnsupervisedSkillDiscovery2022,zengAPDLearningDiverse2022,shafiullahOneAnotherLearning2022,choUnsupervisedReinforcementLearning2022}.
Some algorithms relax the requirement of a task-agnostic prior by selecting the most promising skill before finetuning on the task \citep{rhimEfficientTaskAdaptation2022,laskinCICContrastiveIntrinsic2022}.}
All categories have one thing in common: 
They traverse the policy space during pretraining and potentially encounter behavior close to the optimal prior policy for the finetuning tasks.
There are also algorithms that show great performance by combining both branches of \ac{URL} by simultaneously learning the dynamics and possible behaviors in a domain \revt{\citep{sharmaDynamicsawareUnsupervisedDiscovery2020,yuanEUCLIDEfficientUnsupervised2022}}.
In this work, however, we focus on model-free approaches. 
\section{Method}
\label{sec:method}

In the following, we first discuss the motivation of our approach by linking it to the theoretical results of \citet{eysenbachInformationGeometryUnsupervised2022}.
Based on that, we derive the regularization strategy of \polter using a mixture ensemble policy and discuss the concrete implementation of our method. 
To ground our theoretical motivation, we show exemplary behaviour of \revteditor{\ac{URL} algorithms with and without} \polter on simple control tasks.

\subsection{Motivation}
\label{sec:motivation}

\revteditor{The adaptation objective by \citet{eysenbachInformationGeometryUnsupervised2022}} connects the policy prior after pretraining \revt{$\pi(s)$} with the finetuned policy that has been adapted to an arbitrary downstream task \revt{$\pi^*_T(s)$}.
\revt{As the reward function can be identified with a state-occupancy measure and because there are multiple policies that induce the same state-occupancy measure, the adaptation objective is defined in terms of these state-marginals}:
\begin{equation}
    \min_{\vphantom{\rho^+_T} \rho^*_T(s)\in \mathcal{C}} {\underbrace{\max_{\rho^+_T(s)\in \mathcal{C}} \quad \E_{s \sim \rho^+_T(s)}[r(s)] - \E_{\vphantom{\rho^+_T} s \sim \rho^*_T(s)}[r(s)]}_\text{worst-case regret}}
    + {\underbrace{\infdiv{\vphantom{ \E_{s \sim \rho^+_T(s)} } \rho^*_T(s)}{\rho(s)}}_\text{information cost}}
    \label{eq:adaptation_objective}
\end{equation}
The first term in \cref{eq:adaptation_objective} represents the regret that a \revt{finetuned} policy $\pi^*_T(s)$ achieves with respect to an oracle policy $\pi^+_T(s)$ \revt{for the given task $T$}.
The second term defines the information cost between $\pi^*_T(s)$ and \revt{the prior after pretraining} $\pi(s)$ via their induced state-marginals $\rho^*_T(s)$ and $\rho(s)$.
The set of all feasible state-marginals in an environment is denoted as $\mathcal{C}$.

\begin{rev}
Competence-based \ac{URL} algorithms learn a set of skills $z \in \mathcal{Z}$ during pretraining by maximizing the mutual information $I(\rho(s); z)$ between these skills and the state-marginal.\footnote{\revt{In practice, this is achieved by conditioning the policy on a vector that represents the skill.}} We use the non-standard notation $I(\rho(\cdot); \cdot)$ to differentiate between the mutual information of different state-marginals.
For these algorithms, \citeauthor{eysenbachInformationGeometryUnsupervised2022} demonstrated that the state-marginal averaged over all skills minimizes the distance to the furthest skill. Note, that this skill is generally unknown and depends on the geometric structure of the state-marginal space of the domain:

\begin{lemma}[Lemma 6.5 in \citet{eysenbachInformationGeometryUnsupervised2022}, Theorem \revteditor{13.1.1} in \citet{coverElementsInformationTheory2006}]
    Let $\rho(s\mid z)$ be given. Maximizing the mutual information is equivalent to minimizing the divergence between the average state-marginal $\rho(s)$ and the state-marginal of the furthest possible skill $\revteditor{z \in \mathcal{Z}}$:

    \begin{equation}
        \max_{p(z)} I \left(\rho(s); z\right)= \min_{\rho(s)} \max_{z} \infdiv{\rho(s \mid z)}{\rho(s)}
    \end{equation}
    \label{theorem:average_skill}
\end{lemma}
Thus, the average state-marginal inducing policy minimizes the worst-case information cost of adapting the policy after pretraining to the downstream task $T$.
Competence-based algorithms estimate this optimal prior policy by conditioning their policy on the average of the skill space that they have learned during pretraining.\footnote{\revt{For example, if the skill space $\mathcal{Z}$ lies within a unit hypercube, the vector in the center of the hypercube is chosen.}}
\end{rev}

Data- and knowledge-based algorithms, however, do not try to estimate the optimal prior and thus perform worse in the \ac{URL} setting \citep{laskinCICContrastiveIntrinsic2022}.
Additionally, these algorithms explore the policy space of the domain on a random trajectory that might be close to the optimal prior at one point, but can deviate from it arbitrarily everywhere else.
These disadvantages motivate our key idea to improve the sample-efficiency of data- and knowledge-based \ac{URL} algorithms.

\subsection{\acsu{POLTER}: \acl{POLTER}}
\label{sec:polter_derivation}

\begin{rev}
In this section, we adapt the results from \cref{sec:motivation} for data- and knowledge-based \ac{URL} algorithms and derive our regularization \polter.
The mutual information in \cref{theorem:average_skill} can be decomposed in the entropy of the state-marginal and a conditional entropy term:
\begin{equation}
    I\left(\rho(s); z\right) = H\left(\rho(s)\right) - H\left(p(\rho(s) \mid z)\right)
    \label{eq:mutual_information}
\end{equation}
Data-based algorithms maximize the first term in \cref{eq:mutual_information}, i.e. they maximize an upper bound of the mutual information by maximizing the entropy of the state-marginal distribution.
For knowledge-based algorithms this connection is not as explicit.
However, our experiments in \cref{sec:entropy_polter} show that they also achieve a similar state-marginal entropy.
\begin{reveditor}
Both categories of \ac{URL} algorithms use a mechanism $\mathcal{M}$ (e.g. the prediction error of a dynamics model) that encodes their objective as an intrinsic reward that the \ac{RL} agent is trying to maximize during pretraining.
\end{reveditor}
To apply \cref{theorem:average_skill} to these algorithms, we have to show that the conditional entropy vanishes.

\begin{wrapfigure}{r}{0.5\textwidth}
    \centering
    \includegraphics[width=0.8\linewidth]{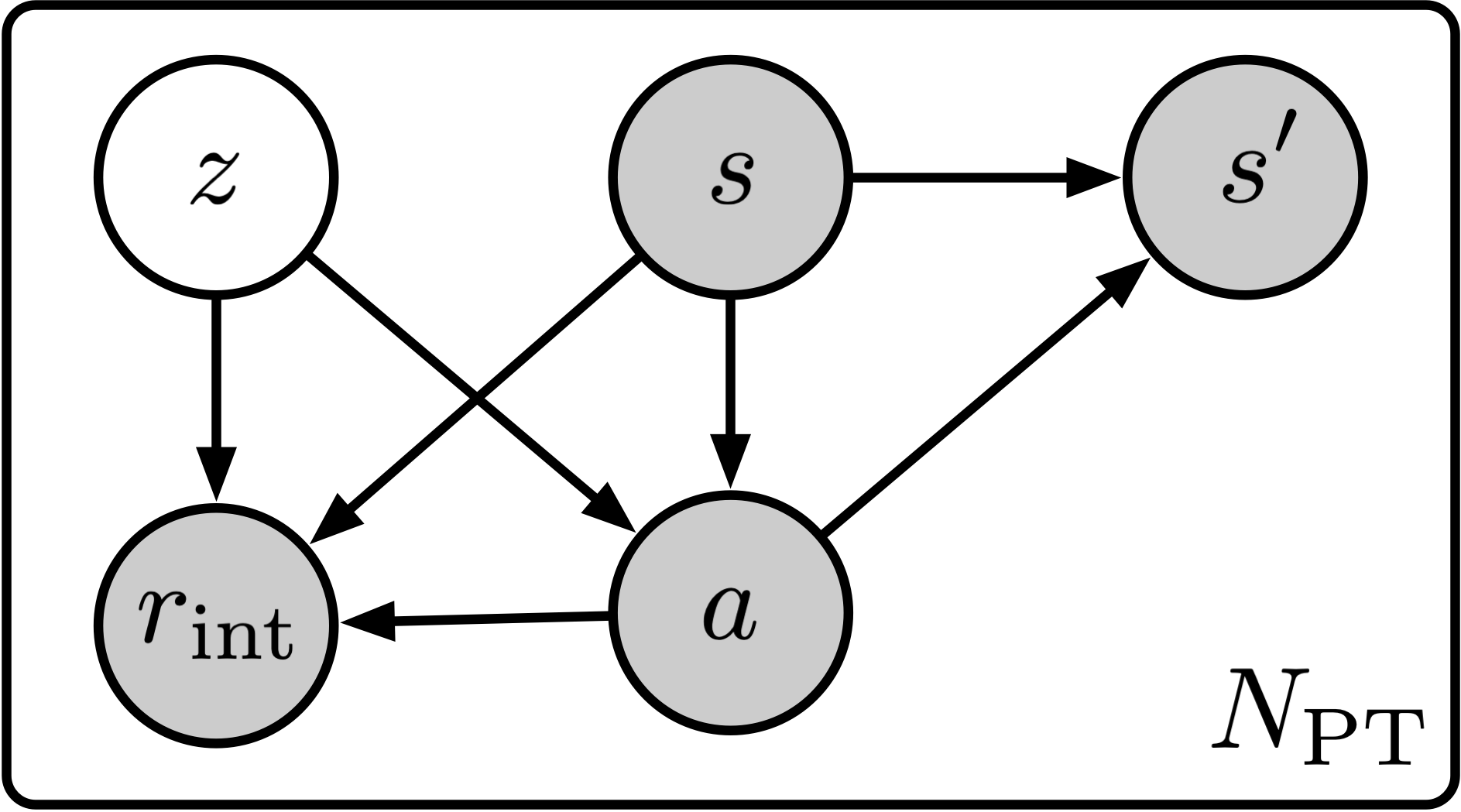}
    \caption{Probabilistic graphical model that underlies the assumption of \polter for approximating the optimal policy using an ensemble of pretraining policies. The latent skill $z$ determines the intrinsic return $r_{\text{int}}$ via the pretraining objective, and the actions $a$ of the agent's policy are implicitly conditioned on it.}
    \label{fig:pgm_polter}
\end{wrapfigure}

Our assumptions can be expressed as the latent variable model in~\cref{fig:pgm_polter} where everything is observed except for the skill $z$ which determines the intrinsic return $r_{\text{int}}$.
At different points during the pretraining steps $0 \leq t < N_{\text{PT}}$, we observe different intrinsic returns for taking an action $a$ in a state $s$. 
Thus, the underlying skill that the agent is supposed to learn must have changed.
This leads us to the following result:
\begin{proposition}
    Let a mechanism \revteditor{$\mathcal{M}$} be given that provides an intrinsic return.
    At a fixed point during pretraining, this mechanism defines a reward function $r_i(s)$. \revteditor{Each reward function can be identified with a latent skill $z_i$, i.e.~there exists a mapping between the set of reward functions and the set of skills}.
    Let a policy $\pi_i(s)$ with state-marginal $\rho_i(s)$ that maximizes this reward function be given.
    Then the conditional entropy of the optimal state-marginal and the skill $H\left(p(\rho_i(s) \mid z_i)\right)$ is equal to 0.
    \label{theorem:optimal_state_marginal}
\end{proposition}
\begin{hproof}
To maximize the intrinsic return for a given skill $z_i$, the optimal policy $\pi_i(s)$ has to place all of the probability mass on the optimal state-marginal, i.e.
$H\left(p(\rho_i(s) \mid z_i)\right) = \E [-\log p(\rho_i(s) \mid z_i)] = \E [-\log 1] = 0$.
If there is a policy that induces a different conditional state-marginal distribution with higher return, then the policy $\pi_i(s)$ is not optimal and there is a contradiction.
\end{hproof}

Thus, under \cref{theorem:optimal_state_marginal} and given an agent that learns the optimal policy for the current state of the intrinsic reward mechanism \revteditor{$\mathcal{M}$}, the conditional entropy in \cref{eq:mutual_information} is minimized.
\revteditor{Given} these assumptions, the data- and knowledge-based \ac{URL} algorithms also maximize the mutual information between the state-marginal and the latent skill and we can apply \cref{theorem:average_skill} \revteditor{to them}.
This suggests that estimating the policy which induces the average state distribution is possible by selecting policies during pretraining that maximize different reward functions, i.e.~learn different latent skills.
\end{rev}
Therefore, we can approximate the optimal \revteditor{prior} policy by an ensemble consisting of a set of policies from different points of pretraining.

\paragraph{Implementation}
The policy loss $\mathcal{L}^{\text{URL}}$ of the \ac{RL} agent is augmented by our approach \polter{} with the following regularization term:
\begin{equation}\label{eq:polter_loss}
    \mathcal{L}^{\text{\polter}}(\pi) = \mathcal{L}^{\text{URL}}(\pi) + \alpha \infdiv{\tilde{\pi}}{\pi}
\end{equation}
\begin{rev}
where $\tilde{\pi}(s) = \sum_k \phi_k \pi_k(s)$ is the mixture ensemble policy consisting of $k$ previous policies from the URL trajectory.
The mixture components $\phi_k = \frac{1}{k}$ parametrize a categorical distribution to construct the non-markovian average ensemble policy. At the start of each episode, the ensemble policy samples the index of an ensemble member with probability $\frac{1}{k}$ and follows this policy afterwards. In the limit, this ensemble policy induces the average state-marginal (see Appendix A.2 in \citet{eysenbachInformationGeometryUnsupervised2022}).
\end{rev}

\begin{algorithm}[ht!]
\caption{\texttt{\ac{URL}\textcolor{green}{+POLTER}}}\label{alg:polter}
\begin{algorithmic}[1]
\footnotesize
\Require Initialize \ac{URL} algorithm, policy $\pi_{\theta, 0}$, replay buffer $\mathcal{D}$, pretraining steps $N_{\mathrm{PT}}$ \Comment \ac{URL} init
\Require \textcolor{green}{Empty ensemble $E = \emptyset$, ensemble snapshot time steps $\mathcal{T}_E$, regularization weight $\alpha$} \Comment \polter init
\For{$t = 0 \dots N_{\mathrm{PT}} - 1$} \Comment{Unsupervised pretraining}
    \If{Beginning of episode}
        \State Observe initial state $s_t \sim p_0(s_0)$
        \If{\textcolor{green}{$t \in \mathcal{T}_E$}} \Comment Update ensemble policy
            \State \textcolor{green}{Extend ensemble $E \leftarrow E \cup \pi_{\theta, t}$} \label{lst:line:add_ensemble}
            \State \textcolor{green}{Update ensemble policy $\tilde\pi$}
        \EndIf
    \EndIf
    \State Choose action $a_t \leftarrow \pi_{\theta, t}(a_t \mid s_t)$
    \State Observe next state $s_{t + 1} \sim p(s_{t + 1} \mid s_{t}, a_t)$ 
    \State Add transition to replay buffer $\mathcal{D} \leftarrow \mathcal{D} \cup (s_t,  a_t, s_{t+1})$
    \State Sample a minibatch $B \sim \mathcal{D}$
    \State Compute loss $\mathcal{L}^\text{\polter} = \mathcal{L}^{\text{URL}}(\pi_{\theta, t}) \textcolor{green}{+\alpha \infdiv{\tilde{\pi}}{\pi_{\theta, t}}}$ \label{lst:line:loss}
    \State Update policy $\pi_{\theta, t}$ with $\mathcal{L}^\text{\polter}$
\EndFor
\State \dots \Comment{Supervised finetuning on task T}
\end{algorithmic}
\end{algorithm} 

In \cref{fig:motivation}, we illustrate the progression of the optimization trajectory in the induced state-marginal space $\rho(s)$ of a policy $\pi(s)$ under the \ac{URL} objective with \polter regularization.
Along its trajectory, the policy $\pi(s)$ is attracted to the ensemble policy $\tilde\pi$ and, thus, closer to the optimal prior $\pi_T^*$ for a specific finetuning task $T$.
This results in a reduced information cost to adapt the pretraining policy to the task and improves the sample-efficiency of the regularized \ac{URL} algorithm.

We outline the combination of \ac{URL} with our regularization \polter in \cref{alg:polter}.
We fix the ensemble policy to a finite set of states to stabilize training, i.e., each time we add a member to the ensemble (\cref{lst:line:add_ensemble}), we condition the ensemble policy on the next state.
Finally, we compute the \ac{KL-divergence} loss and add it to the policy loss of the \ac{URL} algorithm (\cref{lst:line:loss}).
The memory overhead is dominated by the requirement to store the ensemble policies.
For the computational overhead, we only have to consider the $k$ forward passes for conditioning the ensemble policy on a state when adding a new member.\footnote{In our experiments, each policy takes neglected amount of around $4$MB of storage and the wall-clock time increases by 12\%.} 

\subsection{Demonstration on PointMass and Pendulum}
\label{sec:pointmass_demo}

\paragraph{PointMass}
To ground our theoretical motivation in empirical evidence, we evaluate the effect of our regularization on \ac{RND}~\citep{burdaExplorationRandomNetwork2019}, a knowledge-based \ac{URL} algorithm, and \ac{ProtoRL}~\citep{yaratsReinforcementLearningPrototypical2021}, a data-based \ac{URL} algorithm, in the simplistic PointMass environment~\citep{tassaDeepMindControlSuite2018,tassaDmControlSoftware2020}.
In PointMass, we can define the optimal prior $\pi^*_T(s)$ that maximizes the average performance for the downstream finetuning tasks.
This optimal prior policy moves the ball into the center of the area, which minimizes the distance to any possible target location that represent a finetuning task.
With the optimal prior, we can compute the \ac{KL-divergence} between the optimal prior policy and the current pretraining policy $\infdiv{\pi^*_T}{\pi}$.
With \polter, the \ac{KL-divergence} to the optimal prior policy $\pi^*_T$ is generally lower throughout the whole pretraining phase.

\begin{figure}[!ht]
    \centering
    \begin{subfigure}[t]{0.49\textwidth}
         \centering
         \includegraphics[width=\textwidth]{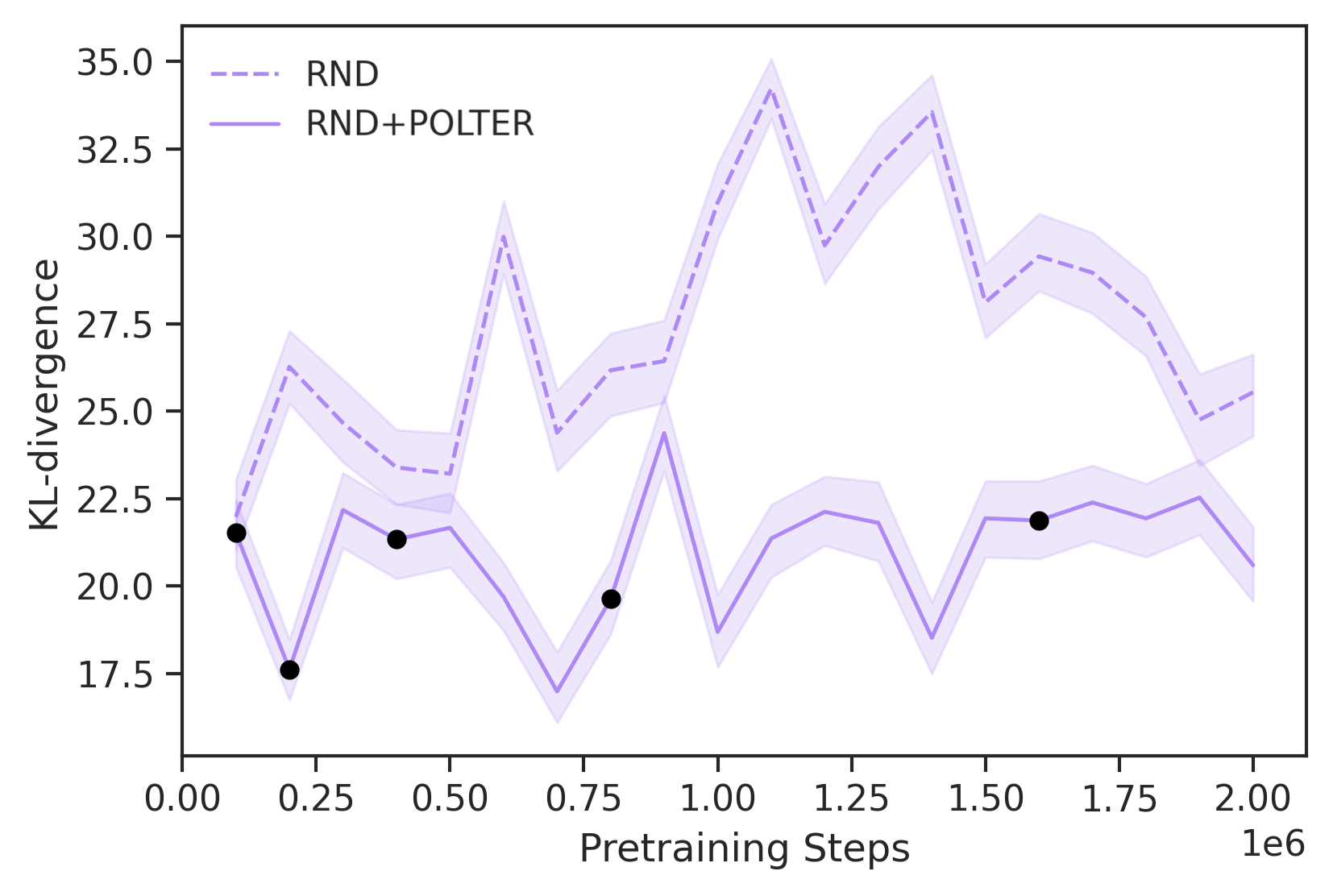}
         \caption{\ac{RND}}
         \label{fig:pointmass_kl}
     \end{subfigure}
     \begin{subfigure}[t]{0.49\textwidth}
         \centering
         \includegraphics[width=\textwidth]{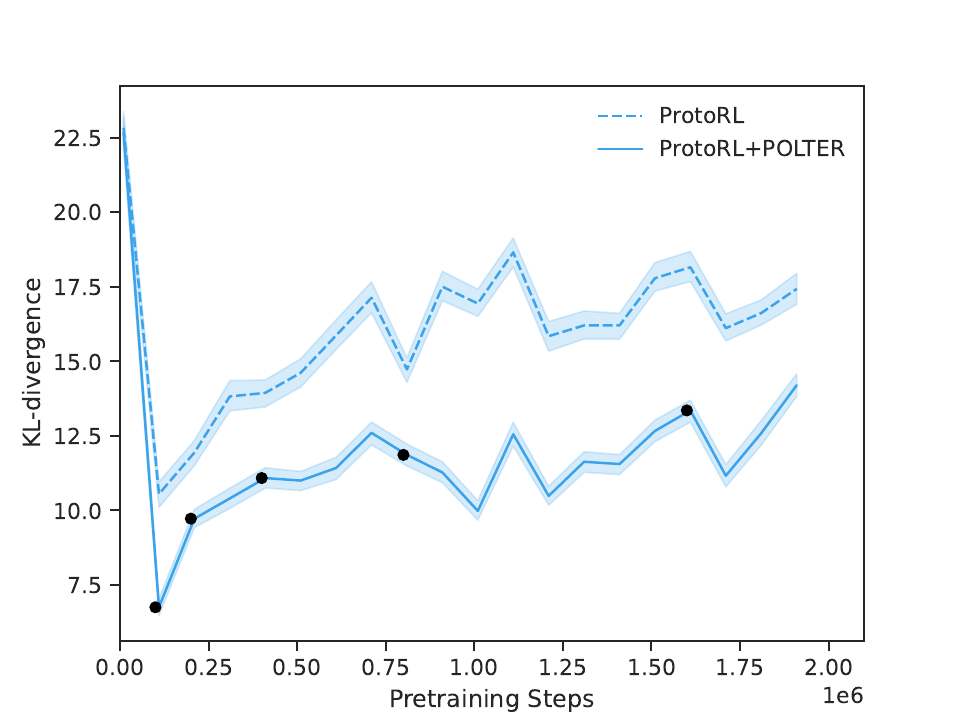}
         \caption{\ac{ProtoRL}}
         \label{fig:pointmass_proto}
     \end{subfigure}
    \caption{Average \ac{KL-divergence} of \ac{RND}/\ac{ProtoRL} (dashed) and \ac{RND}+\polter/\ac{ProtoRL}+\polter (solid) between the pretraining policy $\pi(s_0)$ and the optimal pretraining policy $\pi^*_T(s_0)$ in the PointMass domain during reward-free pretraining. The shaded area indicates the standard error over 10 seeds and the dots are the pretraining checkpoints that were used for the ensemble.}
\end{figure}
\newpage
\paragraph{Pendulum}

\begin{figure}[!ht]
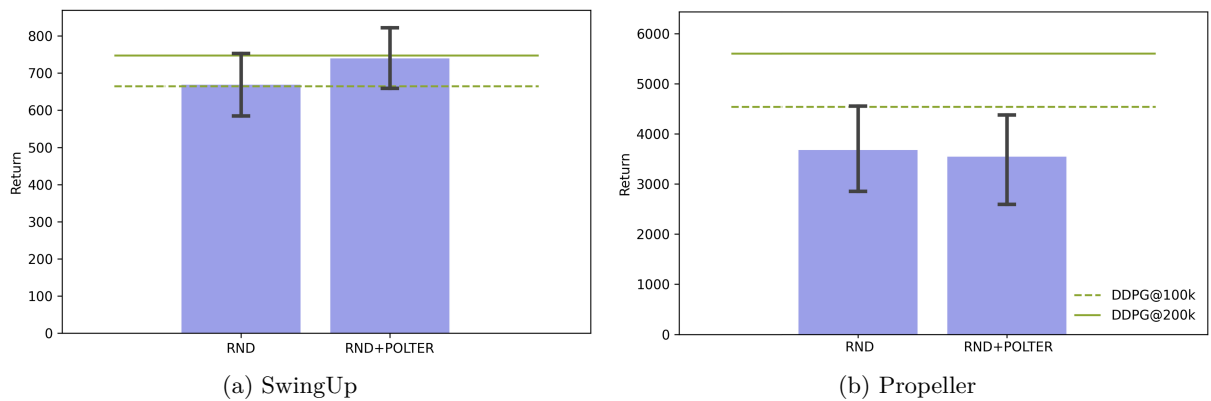

    \centering
    \begin{subfigure}[b]{0.49\textwidth}
         \centering
         \includegraphics[width=\textwidth]{figures/pendulum_rnd_polter_swingup.pdf}
         \caption{SwingUp}
         \label{fig:pendulum_rnd_swingup}
     \end{subfigure}
     \begin{subfigure}[b]{0.49\textwidth}
         \centering
         \includegraphics[width=\textwidth]{figures/pendulum_rnd_polter_propeller.pdf}
         \caption{Propeller}
         \label{fig:pendulum_rnd_propeller}
     \end{subfigure}
\caption{Finetuning performance of \ac{RND} and \ac{RND}+\polter in the Pendulum domain averaged over 10 seeds after $\SI{100}{\kilo\nothing}$ of pretraining. The baseline of \ac{DDPG} is trained for $\SI{100}{\kilo\nothing}$ or $\SI{200}{\kilo\nothing}$ steps without pretraining. The error bars indicate the standard error over 10 seeds.}
    \label{fig:pendulum_rnd_results}
\end{figure}

As described in \cref{fig:motivation}, the effect of \polter on the performance on a specific finetuning task in a given domain depends on the optimal policy for the given task.
Thus, even though \polter improves the average performance over all possible tasks, it might have no or even a detrimental effect on single downstream tasks.
To test this hypothesis, we train \polter on the Pendulum domain for two different downstream tasks.
The first one is the known \emph{SwingUp} task, where the agent has to balance the pole in an upright position.
The second task is a \emph{Propeller} task, which rewards the maximization of the angular velocity of the pole.
This task's optimal policy induces a state-marginal that is further away from the average, which is reflected in \cref{fig:pendulum_rnd_results} in a slightly decreased performance when applying \polter to \ac{RND}.
However, \polter consistently increases the performance on the regular \emph{SwingUp} task.
Further results \revt{for \ac{CIC} \citep{laskinCICContrastiveIntrinsic2022} and \ac{ProtoRL} \citep{yaratsReinforcementLearningPrototypical2021} in this experiment can be found in ~\cref{sec:pendulum_demo}}. 

\section{Experiments}
\label{sec:experimental_setup}

With our theoretical motivation and our prospect on the toy environments PointMass and Pendulum, we now address the following empirical questions at a larger scale: 
\textbf{(Q1)} How does \polter affect the performance after pretraining, and does the performance differ across \ac{URL} algorithm categories?
\textbf{(Q2)} Does \polter improve the sample-efficiency for finetuning?
\textbf{(Q3)} How does the performance vary with an increasing number of pretraining steps? 
\textbf{(Q4)} How does the strength of the regularization affect the performance of different algorithms in different domains?
\revt{\textbf{(Q5)} How to select the checkpoints for the ensemble policy in \polter?}
\revt{\textbf{(Q6)} Do the results also hold for pixel-based observations?}

To answer these questions, we evaluate several different \ac{URL} algorithms with and without our \polter regularization on the \acl{URLB} \citep{laskinURLBUnsupervisedReinforcement2021}.
We use the provided code from \citet{laskinURLBUnsupervisedReinforcement2021} to aid reproducibility and provide our source code in the supplementary material. 
Knowledge-based algorithms, such as \ac{RND} \citep{burdaExplorationRandomNetwork2019}, \acsu{Disagreement} \citep{pathakSelfSupervisedExplorationDisagreement2019}, and \ac{ICM} \citep{pathakCuriositydrivenExplorationSelfsupervised2017}, use the norm or variance of their prediction errors of some aspect of the environment as a learning signal.
The second category consisting of \ac{APT} \citep{schwarzerPretrainingRepresentationsDataEfficient2021}, and \ac{ProtoRL} \citep{yaratsReinforcementLearningPrototypical2021}, are data-based algorithms, that try to maximize the entropy of the state-visitation distribution to explore the environment.
The last category is competence-based algorithms that try to learn a set of skills by maximizing the mutual information between a learned skill vector and some features of the observed states.
Besides the three algorithms that were used by \citet{laskinURLBUnsupervisedReinforcement2021}, \ac{APS} \citep{liuAPSActivePretraining2021}, \ac{SMM} \citep{leeEfficientExplorationState2020}, and \ac{DIAYN} \citep{eysenbachDiversityAllYou2019}, we evaluate our method with a newer method from this category, called \ac{CIC} \citep{laskinCICContrastiveIntrinsic2022}. 

\begin{figure*}[t!]
    \centering
    \includegraphics[width=0.9\textwidth]{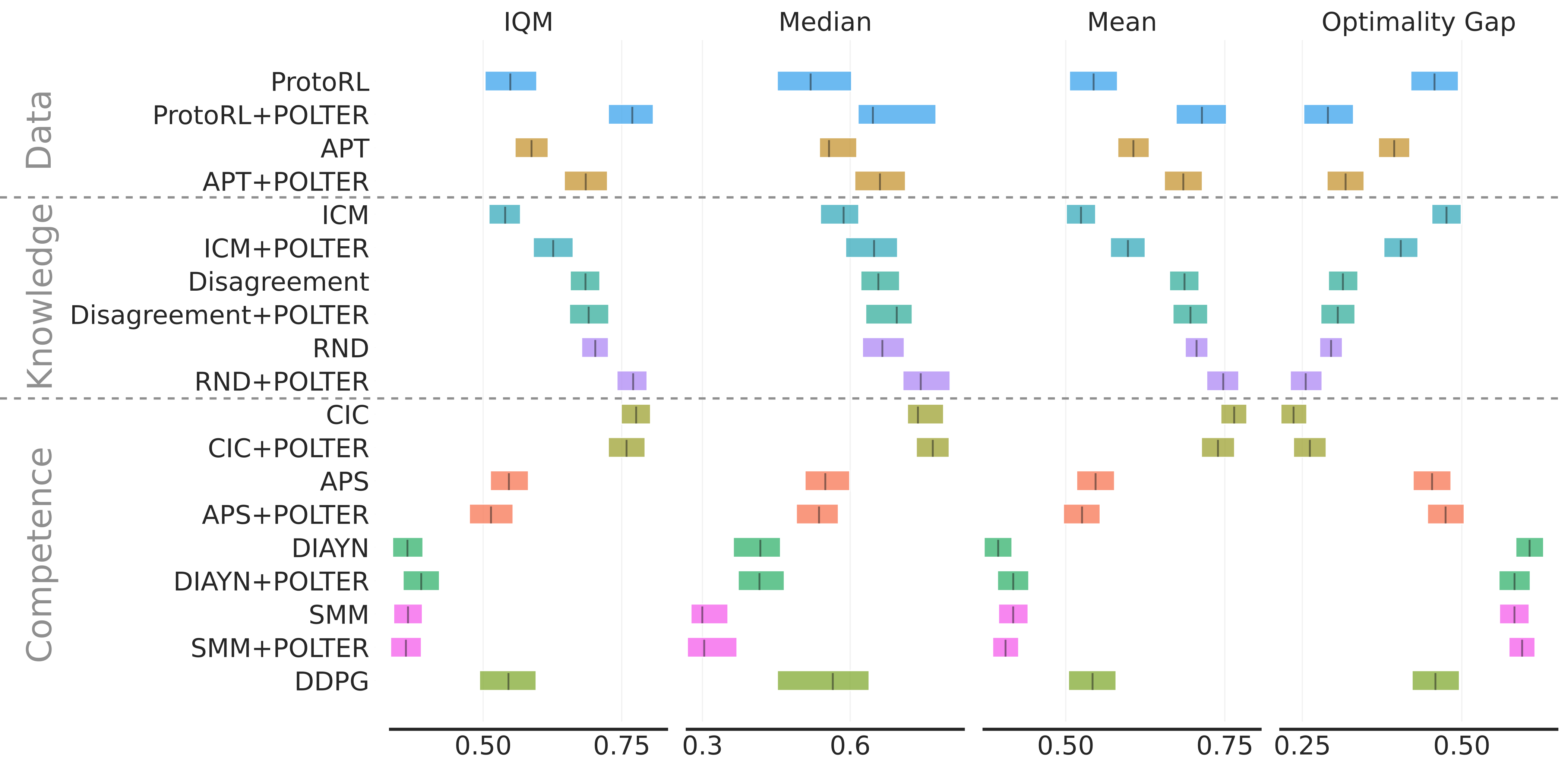}
    \caption{Aggregate statistics of applying \polter to several \ac{URL} algorithms after pretraining for $\SI{2}{\mega\nothing}$ steps. \polter improves the performance of most algorithms substantially, for \acsu{ProtoRL} even by 40\%. Each algorithm is tested on the \ac{URLB} with its 12 tasks from 3 domains using 10 independent seeds, resulting in 120 runs per algorithm. The error bars indicate the 95\% bootstrap confidence intervals.
    \ac{DDPG} is the baseline without pretraining.}
    \label{fig:urlb_metrics}
\end{figure*} 
\section{Results}
\label{sec:results}

\paragraph{Evaluation}

For evaluation, the agent performs pretraining in a reward-free setting for a total of $N_{PT}=\SI{2}{\mega\nothing}$ steps in one of the three domains using one of the described algorithms with or without our \polter regularization.
The performance is then measured as the average return over \num{10} episodes after finetuning for $\SI{100}{\kilo\nothing}$ steps given external rewards that describe the downstream task.
We follow the exact evaluation protocol of the \ac{URLB} and use \ac{DDPG} \citep{lillicrapContinuousControlDeep2016} on state-based observations.
For its hyperparameters and the setup of \polter, see \cref{sec:hyperparameters}.
We follow \citet{agarwalDeepReinforcementLearning2021} in their evaluation using the \ac{IQM} as our main metric\footnote{Although we used the provided code by \citet{laskinURLBUnsupervisedReinforcement2021,laskinCICContrastiveIntrinsic2022} for all baselines, our reproduced results slightly differ from their reported performance.}.
The expert performance is provided by~\citet{laskinURLBUnsupervisedReinforcement2021}.
Each algorithm is tested on 12 tasks from 3 domains using 10 independent seeds, resulting in 120 runs per algorithm. 

\subsection*{Q1: How does \polter affect the performance of different algorithm categories?}

We report the finetuning performance of each \ac{URL} algorithm after $\SI{2}{\mega\nothing}$ pretraining steps with and without \polter in \cref{fig:urlb_metrics}.
With our method, we see an increase in performance for data- and knowledge-based algorithms by $19\%$ \ac{IQM} on average.
For the data-based method \ac{ProtoRL} pretraining with \polter increases the performance even by \num{0.22} \ac{IQM} (\num{40}\%).
\revt{
This result supports the argument in \cref{sec:polter_derivation}, as both data-based algorithms directly maximize the entropy of the state-marginal distribution and with it the upper bound of the mutual information between the state-marginal and the latent skill.
}
The improvement due to \polter also depends on the exploration capabilities of the \ac{URL} algorithm.
This can be seen for \ac{Disagreement}+\polter, which performs worst of all knowledge-based algorithms and is due to the limited exploration of the state space, as has recently been shown by \citet{yaratsDonChangeAlgorithm2022}.
As noted by \citet{laskinURLBUnsupervisedReinforcement2021}, the baseline performances notably differ among themselves, and most competence-based methods are worse than the supervised \ac{DDPG} baseline without pretraining.
The application of \polter has no noticeable effect on their overall performance.
This is due to the similar effect of their skill-averaging and the ensemble, which both pull the policy to the optimal prior.
However, in \cref{sec:analysis_reg_strength}, we show that \polter can be beneficial even for these algorithms.
For more detailed results see \cref{sec:detailed_results_urlb} and \cref{sec:polter_pixel_urlb}.
\begin{figure}[ht!]
    \centering
    \includegraphics[width=0.9\linewidth]{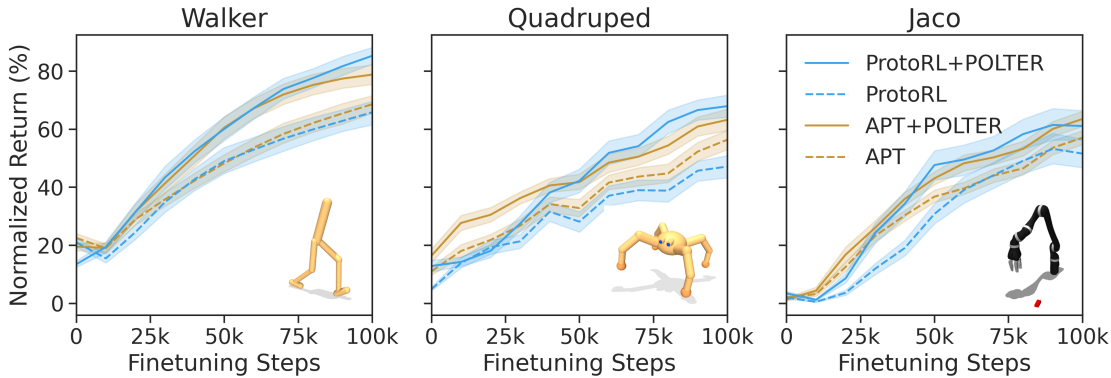}
    \caption{Data-based \ac{URL} algorithms with (solid) and without \polter (dashed) regularization after $\SI{2}{\mega\nothing}$ steps of pretraining. The shaded area indicates the standard error. \polter speeds up finetuning by $\approx 40\%$ on Walker and Quadruped and $\approx 10\%$ on Jaco.}
    \label{fig:finetuning_improvement}
\end{figure} 
\subsection*{Q2: Does \polter improve the sample-efficiency for finetuning? }

As described in \cref{sec:method}, we try to reduce the information cost of finetuning a prior policy on a specific task to improve the sample-efficiency.
Because the effect of our method is most pronounced in the data-based algorithm category, in \cref{fig:finetuning_improvement} we show the expert normalized return of finetuning \ac{APT} and \ac{ProtoRL} with and without \polter after $\SI{2}{\mega\nothing}$ steps of pretraining.
The improvement in sample-efficiency is clearly visible, especially in the locomotion domains.
Here, \polter achieves a speed-up of $\approx 40\%$ compared to the unregularized \ac{URL} algorithms.
It should be noted that the initial performance directly after pretraining is similar for all variants, and they all begin to improve after the same number of finetuning steps.
This implies that exploration during finetuning is not the deciding factor for the performance differences. 
\subsection*{Q3: How does the performance vary with an increasing number of pretraining steps?}

\begin{figure}[hb!]
    \centering
    \includegraphics[width=\linewidth]{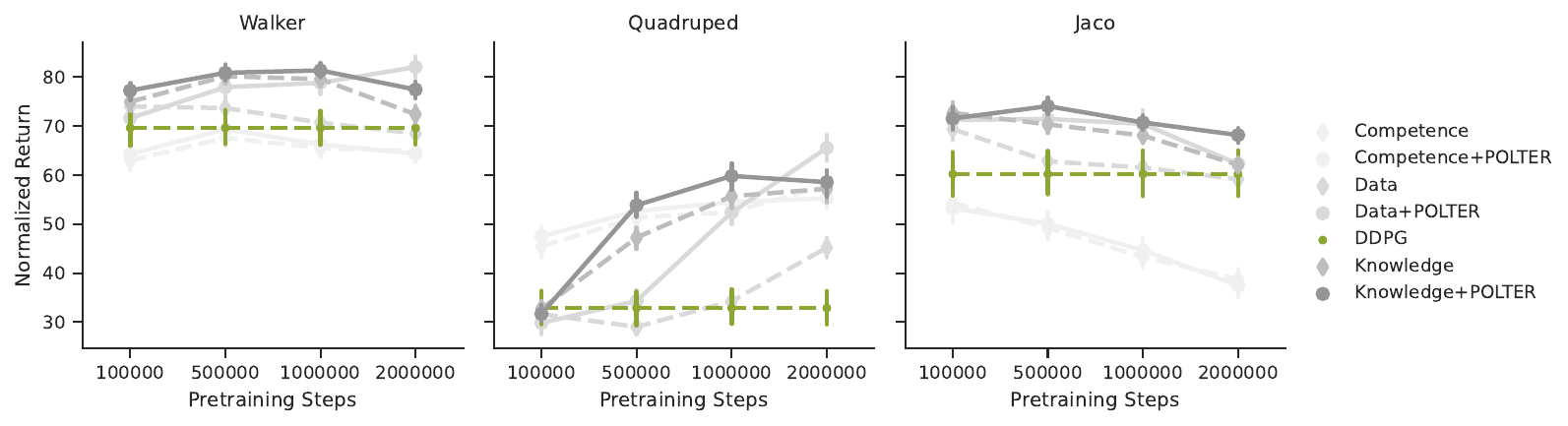}
    \caption{Normalized return per task and algorithm category with (solid) and without (dashed) \polter regularization after finetuning from different pretraining snapshots. \ac{DDPG} is the baseline without pretraining.}
    \label{fig:pretraining_categories}
\end{figure}

In their original publication, \citet{laskinURLBUnsupervisedReinforcement2021} noted that the performance of most \ac{URL} algorithms decreased with an increasing number of pretraining steps.
To evaluate the effect of applying \polter during pretraining, we observe the performance at different numbers of pretraining frames for the three domains of \ac{URLB}.
In \cref{fig:pretraining_categories}, we present our results for knowledge- and data-based methods.
Applying \polter (solid) improves performance at all steps over the baseline (dashed) during pretraining, especially in the locomotion domains.
Jaco is a special case because the arm is in a fixed location, and the return is dominated by the ability to bring the arm into a specific position.
Thus, the domain relies much less on deeper exploration compared to the locomotion environments, and the initial amount of pretraining for  $\SI{100}{\kilo\nothing}$ steps is sufficient to improve on the \ac{DDPG} baseline without pretraining.

\begin{figure*}[!h]
    \centering
    \includegraphics[width=\textwidth]{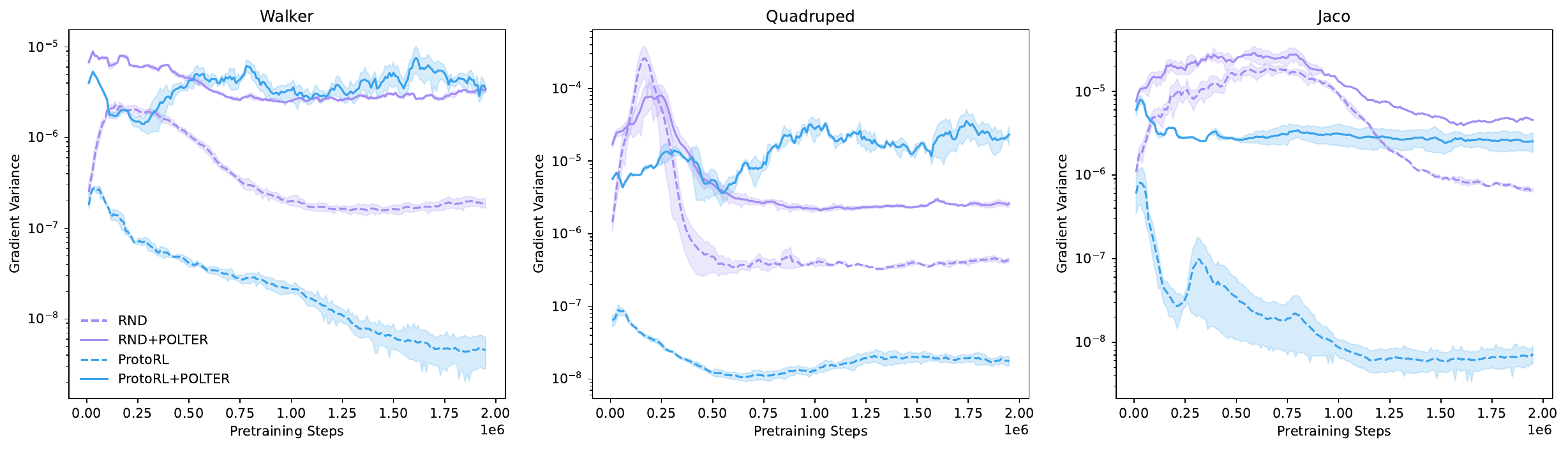}
    \caption{Variance of the actor gradients for \ac{ProtoRL} and \ac{RND} with and without \polter regularization on the \ac{URLB} averaged over \num{10} seeds with a sliding window size of \num{5}. The shaded area indicates the standard error of the mean.}
    \label{fig:gradient_variances}
\end{figure*}

\begin{rev}
The results in \cref{fig:pretraining_categories} also show an increase of the effect of \polter on the downstream performance with an increasing number of pretraining steps.
This leads to the question of what specific effect \polter has on the \ac{URL} algorithm and in particular on the actor network where the regularisation term is applied.
In \cref{fig:gradient_variances}, we computed the variance of the actor gradients over a window of \num{5} gradient steps for \ac{RND} and \ac{ProtoRL} with and without \polter.
The results support the findings of \citeauthor{laskinURLBUnsupervisedReinforcement2021} that longer pretraining is often not beneficial for the downstream performance, as the variance of the gradients and thus, the changes to the network weights decrease substantially for both baselines.
With \polter, however, the variance of the gradients does not decrease as much.
It is also apparent that the difference between \polter and the baseline is much more pronounced for \ac{ProtoRL} than for \ac{RND}.
This result is also in line with the detailed analysis in \cref{fig:pretraining_summary}.
\end{rev}

\subsection*{Q4: How does the strength of the regularization affect the performance?}
\label{sec:analysis_reg_strength}

A hyperparameter of our method is the strength $\alpha$ of our regularization, see \cref{eq:polter_loss}.
We set this to $1.0$ in our preceding evaluations as a natural choice.
To study the sensitivity of this hyperparameter, we evaluated \polter with different settings on a set of representative \ac{URL} algorithms. 
\begin{figure}[ht!]
    \centering
    \includegraphics[width=0.95\linewidth]{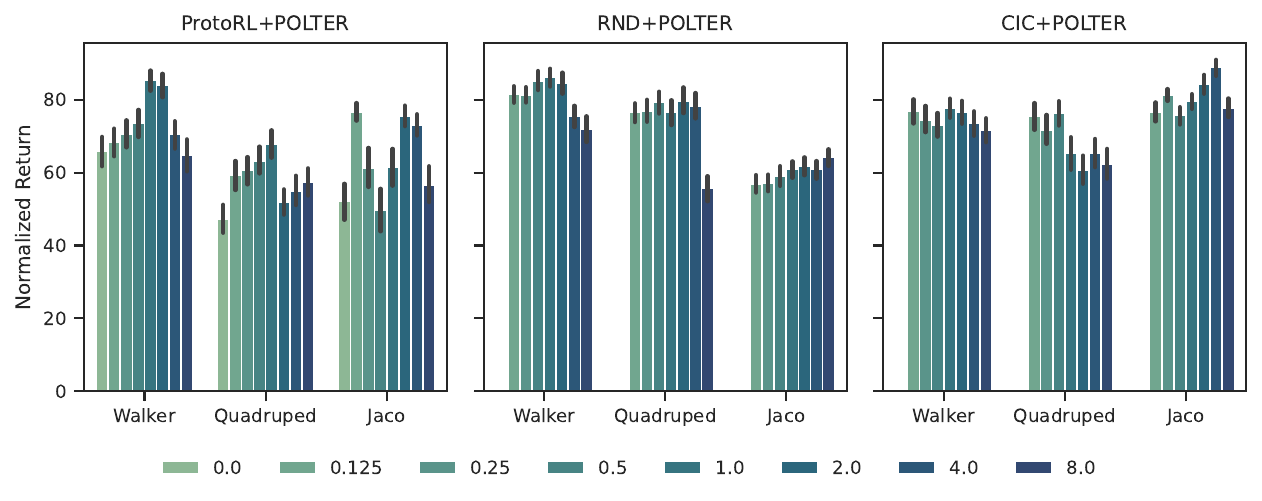}
    \caption{Average normalized return of \ac{ProtoRL}+\polter, \ac{RND}+\polter and \ac{CIC}+\polter after $\SI{2}{\mega\nothing}$ steps of pretraining and $\SI{100}{\kilo\nothing}$ finetuning for different values of the regularization strength $\alpha$.}
    \label{fig:results_alpha_sweep}
\end{figure} 
In \cref{fig:results_alpha_sweep}, we present the performance of the \polter regularized variants of \ac{ProtoRL}, \ac{RND} and \ac{CIC} over different values of $\alpha$.
The results show a robust performance of \polter with its default $\alpha = 1.0$, i.e., \polter performs better or is on par compared to not using \polter ($\alpha= 0.0$).
On average, $\alpha=1.0$ seems to be the best choice.
Nevertheless, the performance of our method can be further improved if it is tuned depending on the task and algorithm.
In the Jaco domain, a higher regularization helps \ac{RND} but might be detrimental for \ac{ProtoRL}.
We hypothesize that this is due to increasingly extreme pretraining policies of \ac{RND} that are dampened via \polter, which is helpful for the tasks in this domain.
For \ac{CIC}, the regularization of \polter has a minor effect or even degrades the performance in the locomotion domains.
As \citet{laskinCICContrastiveIntrinsic2022} noted, locomotion requires a high-entropy pretraining policy that explores the environment well.
\polter is indifferent to the skill vector used by the \ac{CIC} policy in each episode.
So the ensemble contains different skill vectors that are likely to suggest different actions, which partly cancel each other and reduce the exploration.

\begin{figure*}[ht!]
    \centering
    \includegraphics[width=0.95\linewidth]{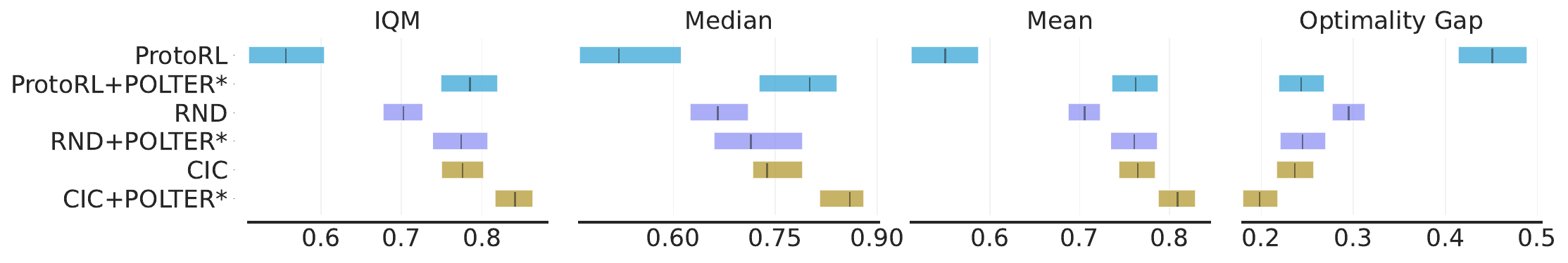}
    \caption{Aggregate statistics of three representative algorithms from each \ac{URL} category after $\SI{2}{\mega\nothing}$ steps of pretraining and $\SI{100}{\kilo\nothing}$ finetuning with tuned regularization strength $\alpha$ in \cref{eq:polter_loss}.}
    \label{fig:results_tuned_iqm}
\end{figure*} 
However, a stronger regularization up to a certain value significantly improves the performance on Jaco, as it does not require much exploration to perform well.
This demonstrates the trade-off between a good prior for finetuning and an attenuated exploration if the regularization is too strong.
It also shows that \polter should be tuned to the domain.

If we apply the insights from \cref{fig:results_alpha_sweep}, we can improve the performance of the three evaluated algorithms even further by tuning $\alpha$ to the locomotion and manipulation domains\footnote{\revt{Tuning here means selecting the best performing $\alpha$ from the previously evaluated grid sweep.}}.
For the results in \cref{fig:results_tuned_iqm}, we set $\alpha$ for \ac{ProtoRL} to $1.0/2.0$ (locomotion/manipulation), for \ac{RND} to $2.0/8.0$ and for \ac{CIC} to $0.0/4.0$. 
We call the tuned variants \ac{ProtoRL}+POLTER*, \ac{RND}+POLTER*\xspace and \ac{CIC}+POLTER*.
Note that our regularization complements the discriminator loss of \ac{CIC}, which is set to $0.9/0.0$ on the two domain categories.
These optimized settings lead to a new state-of-the-art in the \ac{URLB} of $0.84$ \ac{IQM}. 
We note that previous state-of-the-art methods such as \ac{CIC} are also tuned based on the domain of the task and that this experiment allows for a fair comparison.
 
\begin{rev}
\subsection*{Q5: How to select the checkpoints for the ensemble policy in  \polter?}
\label{sec:polter_checkpoints_ablation}

With \polter, we introduce the hyperparameter $\mathcal{T}_E$ that determines which checkpoints should be added to the ensemble.
As described in \cref{sec:polter_setup}, we use the same setting for all our experiments by aligning this hyperparameter with the sign-changes of the second derivative in the intrinsic return during pretraining of \ac{RND} in the Quadruped domain.
Following our derivation in \cref{sec:polter_derivation}, this indicates a change in the latent skill because the actions that the agent takes are now being rated differently by the intrinsic reward mechanism.

\begin{figure}[h!]
    \centering
    \includegraphics[width=\linewidth]{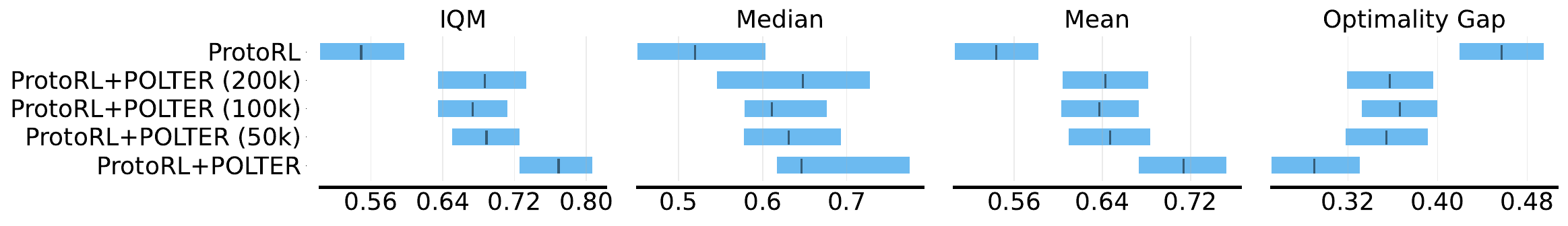}
    \caption{Aggregate statistics on the \ac{URLB} of different linear and logarithmic checkpoint schedules when applying \polter to \ac{ProtoRL}.}
    \label{fig:snapshot_ts_proto_sweep}
\end{figure}

To analyze the sensitivity of our regularization to this hyperparameter, in \cref{fig:snapshot_ts_proto_sweep} we evaluated \ac{ProtoRL} on the \ac{URLB} with three different ensemble schedules.
Each schedule adds a checkpoint every $\{\SI{50}{\kilo\nothing}, \SI{100}{\kilo\nothing}, \SI{200}{\kilo\nothing}\}$ steps.
The results in \cref{fig:snapshot_ts_proto_sweep} indicate that \polter consistently improves the performance of \ac{ProtoRL} for several settings of this hyperparameter.
However, using a logarithmic schedule that aligns with the intrinsic reward dynamics to select the checkpoints further boosts the performance impact of the regularization.
Thus, a better understanding of the latent skill space of the domain and algorithm would enable more sophisticated schedules and increase the performance even further.
\end{rev}

\begin{rev}
\subsection*{Q6: Do the results also hold for pixel-based observations?}
\label{sec:polter_pixel_urlb}

\begin{figure*}[h!]
    \centering
    \includegraphics[width=0.95\textwidth]{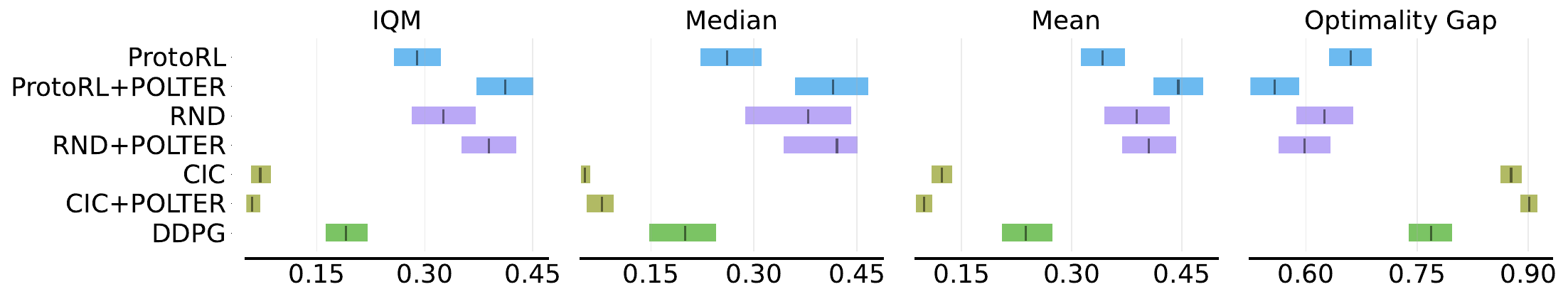}
    \caption{Aggregate statistics of applying \polter to several \ac{URL} algorithms after pretraining for $\SI{2}{\mega\nothing}$ steps. Each algorithm is tested on the \ac{URLB} with its 12 tasks from 3 domains using 10 independent seeds with \textbf{pixel-based observations}, resulting in 120 runs per algorithm. The error bars indicate the 95\% bootstrap confidence intervals.
    \ac{DDPG} is the baseline without pretraining.}
    \label{fig:urlb_metrics_pixels}
\end{figure*} 
\citet{laskinURLBUnsupervisedReinforcement2021} showed that the performance of pixel- and state-based \ac{URL} algorithms differs substantially.
To demonstrate the generality of our approach, we evaluate the effect of \polter on \ac{ProtoRL}, \ac{RND} and \ac{CIC} on the \ac{URLB} with pixel-based observations.
The results in \cref{fig:urlb_metrics_pixels} show that the improvements in \ac{IQM} are approximately the same for \ac{RND} and \ac{ProtoRL} as in state-based environments.
For \ac{CIC} we also see the same effect of a slightly reduced performance.
Note, however, that \ac{CIC} does not work well for the tasks in the Jaco domain.\footnote{\revt{This prompted \citeauthor{laskinCICContrastiveIntrinsic2022} in the original publication to set the weight of their discriminator loss to 0 in the Jaco domain.}}
\end{rev}

\section{Limitations and Future Work}
\label{sec:limitations}

In this work, we focused our experiments on \ac{DDPG} to compare our performance on the \ac{URLB}.
This, however, limits our scope to continuous control problems.
Therefore, extending \ac{URL} to environments with discrete action spaces is a possible next step.
Moreover, a deeper theoretical understanding of the pretraining processes and the connection to the optimal prior policy would enable more sophisticated mixture policies and a better approximation of the optimal prior in the future.
\begin{rev}
As noted in \cref{sec:method}, the space of policies is well explored by the \ac{URL} algorithm, using the average policy is optional under \cref{theorem:optimal_state_marginal}. However, a future research direction is the analysis of the skills that the agent learns during pretraining and incorporate this knowledge into the mixture prior.\footnote{\revt{We explored exponentially decaying mixture policies with \ac{ProtoRL} but found the \ac{IQM} improvement to be slightly worse. For a decay factor of \num{0.9}, \ac{ProtoRL} achieved an \ac{IQM} of \num{0.72} compared to the \num{0.77} of the uniform mixture.}}
\end{rev}
Finally, \ac{POLTER} is not adapted to the category of the \ac{URL} algorithm it is applied to. 
So its performance with competence-based \ac{URL} algorithms leaves room for improvement by taking the observed explicit skills into account. 
\section{Conclusion}
\label{sec:conclusion}

In this work, we introduced \polter (\acl{POLTER}) – a general method to improve the performance of \ac{URL} algorithms.
\ac{URL} is a method to increase the sample-efficiency of \ac{RL} algorithms on a set of tasks by pretraining the policy with an intrinsic reward in the task's domain.
Our regularization pulls the pretraining policy closer to an ensemble of pretraining policies seen during pretraining that correspond to a set of explicit or implicit skills.
We demonstrated the effect of our method in the PointMass and Pendulum domain and extensively evaluated baseline algorithms regularized with \polter on the \ac{URLB} and established its effectiveness in our main experiments.
For data- and knowledge-based \ac{URL} methods, we improved performance on average by $19\%$ and up to $40\%$ (\ac{IQM}).
We argue that our method is a suitable regularization for those algorithms, in contrast to competence-based methods where the effect of \polter is highly algorithm-dependent.
Finally, we showed that our regularization strength can be tuned to the domain and algorithm at hand for further performance improvements, achieving a new state-of-the-art performance with \ac{CIC}+POLTER*. In future work, this manual tuning could be automated utilizing AutoRL~\citep{parker-holderAutomatedReinforcementLearning2022}.
With \polter's easy implementation and negligible computational requirements, we hope it finds its way into more \ac{URL} algorithms and spurs further research on how we can learn general priors for arbitrary tasks.

\section*{Acknowledgements}

Carolin Benjamins and Marius Lindauer acknowledge funding by the German Research Foundation (DFG) under LI 2801/4-1.
Additionally, this work was supported by the Federal Ministry of Education and Research (BMBF), Germany under the project LeibnizKILabor (grant no. 01DD20003) and the AI service center KISSKI (grant no. 01IS22093C), the Center for Digital Innovations (ZDIN) and the Deutsche Forschungsgemeinschaft (DFG) under Germany’s Excellence Strategy within the Cluster of Excellence PhoenixD (EXC 2122). 

\bibliographystyle{plainnat}
\bibliography{tmlr}

\begin{thebibliography}{61}
\providecommand{\natexlab}[1]{#1}
\providecommand{\url}[1]{\texttt{#1}}
\expandafter\ifx\csname urlstyle\endcsname\relax
  \providecommand{\doi}[1]{doi: #1}\else
  \providecommand{\doi}{doi: \begingroup \urlstyle{rm}\Url}\fi

\bibitem[Agarwal et~al.(2021)Agarwal, Schwarzer, Castro, Courville, and Bellemare]{agarwalDeepReinforcementLearning2021}
Rishabh Agarwal, Max Schwarzer, Pablo~Samuel Castro, Aaron~C. Courville, and Marc~G. Bellemare.
\newblock Deep {{Reinforcement Learning}} at the {{Edge}} of the {{Statistical Precipice}}.
\newblock In Marc'Aurelio Ranzato, Alina Beygelzimer, Yann~N. Dauphin, Percy Liang, and Jennifer~Wortman Vaughan, editors, \emph{Advances in {{Neural Information Processing Systems}} 34: {{Annual Conference}} on {{Neural Information Processing Systems}} 2021, {{NeurIPS}} 2021, {{December}} 6-14, 2021, Virtual}, pages 29304--29320, 2021.

\bibitem[Baumli et~al.(2021)Baumli, {Warde-Farley}, Hansen, and Mnih]{baumliRelativeVariationalIntrinsic2021}
Kate Baumli, David {Warde-Farley}, Steven Hansen, and Volodymyr Mnih.
\newblock Relative variational intrinsic control.
\newblock In \emph{Thirty-Fifth {{AAAI}} Conference on Artificial Intelligence, {{AAAI}} 2021, Thirty-Third Conference on Innovative Applications of Artificial Intelligence, {{IAAI}} 2021, the Eleventh Symposium on Educational Advances in Artificial Intelligence, {{EAAI}} 2021, Virtual Event, February 2-9, 2021}, pages 6732--6740. {AAAI Press}, 2021.

\bibitem[Bellemare et~al.(2020)Bellemare, Candido, Castro, Gong, Machado, Moitra, Ponda, and Wang]{bellemareAutonomousNavigationStratospheric2020}
Marc~G. Bellemare, Salvatore Candido, Pablo~Samuel Castro, Jun Gong, Marlos~C. Machado, Subhodeep Moitra, Sameera~S. Ponda, and Ziyu Wang.
\newblock Autonomous navigation of stratospheric balloons using reinforcement learning.
\newblock \emph{Nature}, 588\penalty0 (7836):\penalty0 77--82, December 2020.
\newblock ISSN 0028-0836, 1476-4687.
\newblock \doi{10.1038/s41586-020-2939-8}.

\bibitem[Burda et~al.(2019)Burda, Edwards, Storkey, and Klimov]{burdaExplorationRandomNetwork2019}
Yuri Burda, Harrison Edwards, Amos~J. Storkey, and Oleg Klimov.
\newblock Exploration by random network distillation.
\newblock In \emph{7th {{International Conference}} on {{Learning Representations}}, {{ICLR}} 2019, {{New Orleans}}, {{LA}}, {{USA}}, {{May}} 6-9, 2019}. {OpenReview.net}, 2019.

\bibitem[Chen et~al.(2022{\natexlab{a}})Chen, Gao, Xu, Yang, Li, Ding, Feng, and Wang]{chenNuclearNormMaximization2022}
Chao Chen, Zijian Gao, Kele Xu, Sen Yang, Yiying Li, Bo~Ding, Dawei Feng, and Huaimin Wang.
\newblock Nuclear {{Norm Maximization Based Curiosity-Driven Learning}}, May 2022{\natexlab{a}}.

\bibitem[Chen et~al.(2022{\natexlab{b}})Chen, Wan, Shi, Ding, Gao, and Feng]{chenUncertaintyEstimationBased2022}
Chao Chen, Tianjiao Wan, Peichang Shi, Bo~Ding, Zijian Gao, and Dawei Feng.
\newblock Uncertainty {{Estimation}} based {{Intrinsic Reward For Efficient Reinforcement Learning}}.
\newblock In \emph{2022 {{IEEE International Conference}} on {{Joint Cloud Computing}} ({{JCC}})}, pages 1--8, {Fremont, CA, USA}, August 2022{\natexlab{b}}. {IEEE}.
\newblock ISBN 978-1-66546-285-3.
\newblock \doi{10.1109/JCC56315.2022.00008}.

\bibitem[Cho et~al.(2022)Cho, Kim, and Kim]{choUnsupervisedReinforcementLearning2022}
Daesol Cho, Jigang Kim, and H.~Jin Kim.
\newblock Unsupervised {{Reinforcement Learning}} for {{Transferable Manipulation Skill Discovery}}.
\newblock \emph{IEEE Robotics and Automation Letters}, 7\penalty0 (3):\penalty0 7455--7462, July 2022.
\newblock ISSN 2377-3766, 2377-3774.
\newblock \doi{10.1109/LRA.2022.3171915}.

\bibitem[Cover and Thomas(2006)]{coverElementsInformationTheory2006}
Thomas~M Cover and Joy~A Thomas.
\newblock \emph{Elements of {{Information Theory}}}.
\newblock {Wiley-Interscience}, July 2006.

\bibitem[Degrave et~al.(2022)Degrave, Felici, Buchli, Neunert, Tracey, Carpanese, Ewalds, Hafner, Abdolmaleki, {de las Casas}, Donner, Fritz, Galperti, Huber, Keeling, Tsimpoukelli, Kay, Merle, Moret, Noury, Pesamosca, Pfau, Sauter, Sommariva, Coda, Duval, Fasoli, Kohli, Kavukcuoglu, Hassabis, and Riedmiller]{degraveMagneticControlTokamak2022}
Jonas Degrave, Federico Felici, Jonas Buchli, Michael Neunert, Brendan Tracey, Francesco Carpanese, Timo Ewalds, Roland Hafner, Abbas Abdolmaleki, Diego {de las Casas}, Craig Donner, Leslie Fritz, Cristian Galperti, Andrea Huber, James Keeling, Maria Tsimpoukelli, Jackie Kay, Antoine Merle, Jean-Marc Moret, Seb Noury, Federico Pesamosca, David Pfau, Olivier Sauter, Cristian Sommariva, Stefano Coda, Basil Duval, Ambrogio Fasoli, Pushmeet Kohli, Koray Kavukcuoglu, Demis Hassabis, and Martin Riedmiller.
\newblock Magnetic control of tokamak plasmas through deep reinforcement learning.
\newblock \emph{Nature}, 602\penalty0 (7897):\penalty0 414--419, February 2022.
\newblock ISSN 0028-0836, 1476-4687.
\newblock \doi{10.1038/s41586-021-04301-9}.

\bibitem[Dockhorn and Apeldoorn(2018)]{dockhornForwardModelApproximation2018}
Alexander Dockhorn and Daan Apeldoorn.
\newblock Forward model approximation for general video game learning.
\newblock In \emph{2018 IEEE Conference on Computational Intelligence and Games (CIG)}, pages 1--8, 2018.
\newblock \doi{10.1109/CIG.2018.8490411}.

\bibitem[Ebert et~al.(2018)Ebert, Finn, Dasari, Xie, Lee, and Levine]{ebertVisualForesightModelBased2018}
Frederik Ebert, Chelsea Finn, Sudeep Dasari, Annie Xie, Alex Lee, and Sergey Levine.
\newblock Visual {{Foresight}}: {{Model-Based Deep Reinforcement Learning}} for {{Vision-Based Robotic Control}}.
\newblock \emph{arXiv:1812.00568 [cs]}, December 2018.

\bibitem[Eysenbach et~al.(2019)Eysenbach, Gupta, Ibarz, and Levine]{eysenbachDiversityAllYou2019}
Benjamin Eysenbach, Abhishek Gupta, Julian Ibarz, and Sergey Levine.
\newblock Diversity is {{All You Need}}: {{Learning Skills}} without a {{Reward Function}}.
\newblock In \emph{7th {{International Conference}} on {{Learning Representations}}, {{ICLR}} 2019, {{New Orleans}}, {{LA}}, {{USA}}, {{May}} 6-9, 2019}. {OpenReview.net}, 2019.

\bibitem[Eysenbach et~al.(2022)Eysenbach, Salakhutdinov, and Levine]{eysenbachInformationGeometryUnsupervised2022}
Benjamin Eysenbach, Ruslan Salakhutdinov, and Sergey Levine.
\newblock The {{Information Geometry}} of {{Unsupervised Reinforcement Learning}}.
\newblock In \emph{International {{Conference}} on {{Learning Representations}}}, 2022.

\bibitem[Gregor et~al.(2017)Gregor, Rezende, and Wierstra]{gregorVariationalIntrinsicControl2017}
Karol Gregor, Danilo~Jimenez Rezende, and Daan Wierstra.
\newblock Variational {{Intrinsic Control}}.
\newblock In \emph{5th {{International Conference}} on {{Learning Representations}}, {{ICLR}} 2017, {{Toulon}}, {{France}}, {{April}} 24-26, 2017, {{Workshop Track Proceedings}}}. {OpenReview.net}, 2017.

\bibitem[Guo et~al.(2021)Guo, Azar, Saade, Thakoor, Piot, Pires, Valko, Mesnard, Lattimore, and Munos]{guoGeometricEntropicExploration2021}
Zhaohan~Daniel Guo, Mohammad~Gheshlaghi Azar, Alaa Saade, Shantanu Thakoor, Bilal Piot, Bernardo~Avila Pires, Michal Valko, Thomas Mesnard, Tor Lattimore, and R{\'e}mi Munos.
\newblock Geometric {{Entropic Exploration}}, January 2021.

\bibitem[Ha and Schmidhuber(2018)]{haRecurrentWorldModels2018}
David Ha and J{\"u}rgen Schmidhuber.
\newblock Recurrent {{World Models Facilitate Policy Evolution}}.
\newblock In Samy Bengio, Hanna~M. Wallach, Hugo Larochelle, Kristen Grauman, Nicol{\`o} {Cesa-Bianchi}, and Roman Garnett, editors, \emph{Advances in {{Neural Information Processing Systems}} 31: {{Annual Conference}} on {{Neural Information Processing Systems}} 2018, {{NeurIPS}} 2018, {{December}} 3-8, 2018, {{Montr\'eal}}, {{Canada}}}, pages 2455--2467, 2018.

\bibitem[Hansen et~al.(2022)Hansen, Su, and Wang]{hansenTemporalDifferenceLearning2022}
Nicklas Hansen, Hao Su, and Xiaolong Wang.
\newblock Temporal difference learning for model predictive control.
\newblock In Kamalika Chaudhuri, Stefanie Jegelka, Le~Song, Csaba Szepesv{\'a}ri, Gang Niu, and Sivan Sabato, editors, \emph{International Conference on Machine Learning, {{ICML}} 2022, 17-23 July 2022, Baltimore, Maryland, {{USA}}}, volume 162 of \emph{Proceedings of Machine Learning Research}, pages 8387--8406. {PMLR}, 2022.

\bibitem[Hazan et~al.(2019)Hazan, Kakade, Singh, and Soest]{hazanProvablyEfficientMaximum2019}
Elad Hazan, Sham~M. Kakade, Karan Singh, and Abby~Van Soest.
\newblock Provably {{Efficient Maximum Entropy Exploration}}.
\newblock In Kamalika Chaudhuri and Ruslan Salakhutdinov, editors, \emph{Proceedings of the 36th {{International Conference}} on {{Machine Learning}}, {{ICML}} 2019, 9-15 {{June}} 2019, {{Long Beach}}, {{California}}, {{USA}}}, volume~97 of \emph{Proceedings of {{Machine Learning Research}}}, pages 2681--2691. {PMLR}, 2019.

\bibitem[He et~al.(2022)He, Jiang, Zhang, Shao, and Ji]{heWassersteinUnsupervisedReinforcement2022}
Shuncheng He, Yuhang Jiang, Hongchang Zhang, Jianzhun Shao, and Xiangyang Ji.
\newblock Wasserstein {{Unsupervised Reinforcement Learning}}.
\newblock \emph{Proceedings of the AAAI Conference on Artificial Intelligence}, 36\penalty0 (6):\penalty0 6884--6892, June 2022.
\newblock ISSN 2374-3468.
\newblock \doi{10.1609/aaai.v36i6.20645}.

\bibitem[Jacq et~al.(2022)Jacq, Orsini, {Dulac-Arnold}, Pietquin, Geist, and Bachem]{jacqC3POLearningAchieve2022}
Alexis Jacq, Manu Orsini, Gabriel {Dulac-Arnold}, Olivier Pietquin, Matthieu Geist, and Olivier Bachem.
\newblock {{C3PO}}: {{Learning}} to {{Achieve Arbitrary Goals}} via {{Massively Entropic Pretraining}}, November 2022.

\bibitem[Jiang et~al.(2022)Jiang, Gao, and Chen]{jiangUnsupervisedSkillDiscovery2022}
Zheyuan Jiang, Jingyue Gao, and Jianyu Chen.
\newblock Unsupervised {{Skill Discovery}} via {{Recurrent Skill Training}}.
\newblock In \emph{Advances in {{Neural Information Processing Systems}}}, October 2022.

\bibitem[Jin et~al.(2020)Jin, Krishnamurthy, Simchowitz, and Yu]{jinRewardFreeExplorationReinforcement2020}
Chi Jin, Akshay Krishnamurthy, Max Simchowitz, and Tiancheng Yu.
\newblock Reward-{{Free Exploration}} for {{Reinforcement Learning}}.
\newblock In \emph{Proceedings of the 37th {{International Conference}} on {{Machine Learning}}, {{ICML}} 2020, 13-18 {{July}} 2020, {{Virtual Event}}}, volume 119 of \emph{Proceedings of {{Machine Learning Research}}}, pages 4870--4879. {PMLR}, 2020.

\bibitem[Laskin et~al.(2020)Laskin, Srinivas, and Abbeel]{laskinCURLContrastiveUnsupervised2020}
Michael Laskin, Aravind Srinivas, and Pieter Abbeel.
\newblock {{CURL}}: {{Contrastive Unsupervised Representations}} for {{Reinforcement Learning}}.
\newblock In \emph{Proceedings of the 37th {{International Conference}} on {{Machine Learning}}, {{ICML}} 2020, 13-18 {{July}} 2020, {{Virtual Event}}}, volume 119 of \emph{Proceedings of {{Machine Learning Research}}}, pages 5639--5650. {PMLR}, 2020.

\bibitem[Laskin et~al.(2021)Laskin, Yarats, Liu, Lee, Zhan, Lu, Cang, Pinto, and Abbeel]{laskinURLBUnsupervisedReinforcement2021}
Michael Laskin, Denis Yarats, Hao Liu, Kimin Lee, Albert Zhan, Kevin Lu, Catherine Cang, Lerrel Pinto, and Pieter Abbeel.
\newblock {{URLB}}: {{Unsupervised Reinforcement Learning Benchmark}}.
\newblock In Joaquin Vanschoren and Sai-Kit Yeung, editors, \emph{Proceedings of the {{Neural Information Processing Systems Track}} on {{Datasets}} and {{Benchmarks}} 1, {{NeurIPS Datasets}} and {{Benchmarks}} 2021, {{December}} 2021, Virtual}, 2021.

\bibitem[Laskin et~al.(2022)Laskin, Liu, Peng, Yarats, Rajeswaran, and Abbeel]{laskinCICContrastiveIntrinsic2022}
Michael Laskin, Hao Liu, Xue~Bin Peng, Denis Yarats, Aravind Rajeswaran, and Pieter Abbeel.
\newblock {{CIC}}: {{Contrastive Intrinsic Control}} for {{Unsupervised Skill Discovery}}.
\newblock \emph{arXiv:2202.00161 [cs]}, February 2022.

\bibitem[Lee et~al.(2020)Lee, Eysenbach, Parisotto, Xing, Levine, and Salakhutdinov]{leeEfficientExplorationState2020}
Lisa Lee, Benjamin Eysenbach, Emilio Parisotto, Eric Xing, Sergey Levine, and Ruslan Salakhutdinov.
\newblock Efficient {{Exploration}} via {{State Marginal Matching}}.
\newblock \emph{arXiv:1906.05274 [cs, stat]}, February 2020.

\bibitem[Lillicrap et~al.(2016)Lillicrap, Hunt, Pritzel, Heess, Erez, Tassa, Silver, and Wierstra]{lillicrapContinuousControlDeep2016}
Timothy~P. Lillicrap, Jonathan~J. Hunt, Alexander Pritzel, Nicolas Heess, Tom Erez, Yuval Tassa, David Silver, and Daan Wierstra.
\newblock Continuous control with deep reinforcement learning.
\newblock In Yoshua Bengio and Yann LeCun, editors, \emph{4th {{International Conference}} on {{Learning Representations}}, {{ICLR}} 2016, {{San Juan}}, {{Puerto Rico}}, {{May}} 2-4, 2016, {{Conference Track Proceedings}}}, 2016.

\bibitem[Liu and Abbeel(2021)]{liuAPSActivePretraining2021}
Hao Liu and Pieter Abbeel.
\newblock {{APS}}: {{Active Pretraining}} with {{Successor Features}}.
\newblock In Marina Meila and Tong Zhang, editors, \emph{Proceedings of the 38th {{International Conference}} on {{Machine Learning}}, {{ICML}} 2021, 18-24 {{July}} 2021, {{Virtual Event}}}, volume 139 of \emph{Proceedings of {{Machine Learning Research}}}, pages 6736--6747, 2021.

\bibitem[Mirhoseini et~al.(2021)Mirhoseini, Goldie, Yazgan, Jiang, Songhori, Wang, Lee, Johnson, Pathak, Nazi, Pak, Tong, Srinivasa, Hang, Tuncer, Le, Laudon, Ho, Carpenter, and Dean]{mirhoseiniGraphPlacementMethodology2021}
Azalia Mirhoseini, Anna Goldie, Mustafa Yazgan, Joe~Wenjie Jiang, Ebrahim Songhori, Shen Wang, Young-Joon Lee, Eric Johnson, Omkar Pathak, Azade Nazi, Jiwoo Pak, Andy Tong, Kavya Srinivasa, William Hang, Emre Tuncer, Quoc~V. Le, James Laudon, Richard Ho, Roger Carpenter, and Jeff Dean.
\newblock A graph placement methodology for fast chip design.
\newblock \emph{Nature}, 594\penalty0 (7862):\penalty0 207--212, June 2021.
\newblock ISSN 1476-4687.
\newblock \doi{10.1038/s41586-021-03544-w}.

\bibitem[Mnih et~al.(2015)Mnih, Kavukcuoglu, Silver, Rusu, Veness, Bellemare, Graves, Riedmiller, Fidjeland, Ostrovski, Petersen, Beattie, Sadik, Antonoglou, King, Kumaran, Wierstra, Legg, and Hassabis]{mnihHumanlevelControlDeep2015}
Volodymyr Mnih, Koray Kavukcuoglu, David Silver, Andrei~A. Rusu, Joel Veness, Marc~G. Bellemare, Alex Graves, Martin Riedmiller, Andreas~K. Fidjeland, Georg Ostrovski, Stig Petersen, Charles Beattie, Amir Sadik, Ioannis Antonoglou, Helen King, Dharshan Kumaran, Daan Wierstra, Shane Legg, and Demis Hassabis.
\newblock Human-level control through deep reinforcement learning.
\newblock \emph{Nature}, 518\penalty0 (7540):\penalty0 529--533, February 2015.
\newblock ISSN 0028-0836, 1476-4687.
\newblock \doi{10.1038/nature14236}.

\bibitem[Mohamed and Rezende(2015)]{mohamedVariationalInformationMaximisation2015}
Shakir Mohamed and Danilo~Jimenez Rezende.
\newblock Variational {{Information Maximisation}} for {{Intrinsically Motivated Reinforcement Learning}}.
\newblock In Corinna Cortes, Neil~D. Lawrence, Daniel~D. Lee, Masashi Sugiyama, and Roman Garnett, editors, \emph{Advances in {{Neural Information Processing Systems}} 28: {{Annual Conference}} on {{Neural Information Processing Systems}} 2015, {{December}} 7-12, 2015, {{Montreal}}, {{Quebec}}, {{Canada}}}, pages 2125--2133, 2015.

\bibitem[Mutti et~al.(2021)Mutti, Pratissoli, and Restelli]{muttiTaskAgnosticExplorationPolicy2021}
Mirco Mutti, Lorenzo Pratissoli, and Marcello Restelli.
\newblock Task-{{Agnostic Exploration}} via {{Policy Gradient}} of a {{Non-Parametric State Entropy Estimate}}.
\newblock \emph{Proceedings of the AAAI Conference on Artificial Intelligence}, 35\penalty0 (10):\penalty0 9028--9036, May 2021.
\newblock ISSN 2374-3468, 2159-5399.
\newblock \doi{10.1609/aaai.v35i10.17091}.

\bibitem[Mutti et~al.(2022)Mutti, Mancassola, and Restelli]{muttiUnsupervisedReinforcementLearning2022}
Mirco Mutti, Mattia Mancassola, and Marcello Restelli.
\newblock Unsupervised reinforcement learning in multiple environments.
\newblock In \emph{Thirty-Sixth {{AAAI}} Conference on Artificial Intelligence, {{AAAI}} 2022, Thirty-Fourth Conference on Innovative Applications of Artificial Intelligence, {{IAAI}} 2022, the Twelveth Symposium on Educational Advances in Artificial Intelligence, {{EAAI}} 2022 Virtual Event, February 22 - March 1, 2022}, pages 7850--7858. {AAAI Press}, 2022.

\bibitem[Nedergaard and Cook(2022)]{nedergaardKMeansMaximumEntropy2022}
Alexander Nedergaard and Matthew Cook.
\newblock K-{{Means Maximum Entropy Exploration}}, June 2022.

\bibitem[OpenAI et~al.(2019)OpenAI, Akkaya, Andrychowicz, Chociej, Litwin, McGrew, Petron, Paino, Plappert, Powell, Ribas, Schneider, Tezak, Tworek, Welinder, Weng, Yuan, Zaremba, and Zhang]{openaiSolvingRubikCube2019}
OpenAI, Ilge Akkaya, Marcin Andrychowicz, Maciek Chociej, Mateusz Litwin, Bob McGrew, Arthur Petron, Alex Paino, Matthias Plappert, Glenn Powell, Raphael Ribas, Jonas Schneider, Nikolas Tezak, Jerry Tworek, Peter Welinder, Lilian Weng, Qiming Yuan, Wojciech Zaremba, and Lei Zhang.
\newblock Solving {{Rubik}}'s {{Cube}} with a {{Robot Hand}}.
\newblock \emph{arXiv:1910.07113 [cs, stat]}, October 2019.

\bibitem[Ortega and Lee(2014)]{ortegaAdversarialInterpretationInformationTheoretic2014}
Pedro Ortega and Daniel Lee.
\newblock An {{Adversarial Interpretation}} of {{Information-Theoretic Bounded Rationality}}.
\newblock \emph{Proceedings of the AAAI Conference on Artificial Intelligence}, 28\penalty0 (1), June 2014.
\newblock ISSN 2374-3468, 2159-5399.
\newblock \doi{10.1609/aaai.v28i1.9071}.

\bibitem[Oudeyer et~al.(2007)Oudeyer, Kaplan, and Hafner]{oudeyerIntrinsicMotivationSystems2007}
Pierre-Yves Oudeyer, Frdric Kaplan, and Verena~V. Hafner.
\newblock Intrinsic {{Motivation Systems}} for {{Autonomous Mental Development}}.
\newblock \emph{IEEE Transactions on Evolutionary Computation}, 11\penalty0 (2):\penalty0 265--286, April 2007.
\newblock ISSN 1089-778X.
\newblock \doi{10.1109/TEVC.2006.890271}.

\bibitem[{Parker-Holder} et~al.(2022){Parker-Holder}, Rajan, Song, Biedenkapp, Miao, Eimer, Zhang, Nguyen, Calandra, Faust, Hutter, and Lindauer]{parker-holderAutomatedReinforcementLearning2022}
Jack {Parker-Holder}, Raghu Rajan, Xingyou Song, Andr{\'e} Biedenkapp, Yingjie Miao, Theresa Eimer, Baohe Zhang, Vu~Nguyen, Roberto Calandra, Aleksandra Faust, Frank Hutter, and Marius Lindauer.
\newblock Automated reinforcement learning ({{AutoRL}}): {{A}} survey and open problems.
\newblock \emph{Journal of Artificial Intelligence Research}, 74:\penalty0 517--568, 2022.

\bibitem[Pathak et~al.(2017)Pathak, Agrawal, Efros, and Darrell]{pathakCuriositydrivenExplorationSelfsupervised2017}
Deepak Pathak, Pulkit Agrawal, Alexei~A. Efros, and Trevor Darrell.
\newblock Curiosity-driven {{Exploration}} by {{Self-supervised Prediction}}.
\newblock In Doina Precup and Yee~Whye Teh, editors, \emph{Proceedings of the 34th {{International Conference}} on {{Machine Learning}}, {{ICML}} 2017, {{Sydney}}, {{NSW}}, {{Australia}}, 6-11 {{August}} 2017}, volume~70 of \emph{Proceedings of {{Machine Learning Research}}}, pages 2778--2787. {PMLR}, 2017.

\bibitem[Pathak et~al.(2019)Pathak, Gandhi, and Gupta]{pathakSelfSupervisedExplorationDisagreement2019}
Deepak Pathak, Dhiraj Gandhi, and Abhinav Gupta.
\newblock Self-{{Supervised Exploration}} via {{Disagreement}}.
\newblock In Kamalika Chaudhuri and Ruslan Salakhutdinov, editors, \emph{Proceedings of the 36th {{International Conference}} on {{Machine Learning}}, {{ICML}} 2019, 9-15 {{June}} 2019, {{Long Beach}}, {{California}}, {{USA}}}, volume~97 of \emph{Proceedings of {{Machine Learning Research}}}, pages 5062--5071. {PMLR}, 2019.

\bibitem[Rajeswar et~al.(2022)Rajeswar, Mazzaglia, Verbelen, Pich{\'e}, Dhoedt, Courville, and Lacoste]{rajeswarUnsupervisedModelbasedPretraining2022}
Sai Rajeswar, Pietro Mazzaglia, Tim Verbelen, Alexandre Pich{\'e}, Bart Dhoedt, Aaron Courville, and Alexandre Lacoste.
\newblock Unsupervised model-based pre-training for data-efficient reinforcement learning from pixels.
\newblock In \emph{Decision Awareness in Reinforcement Learning Workshop at {{ICML}} 2022}, 2022.

\bibitem[RHIM et~al.(2022)RHIM, Yang, and Kim]{rhimEfficientTaskAdaptation2022}
{\relax JUNGSUB}~RHIM, Eunseok Yang, and Taesup Kim.
\newblock Efficient task adaptation by mixing discovered skills.
\newblock In \emph{First Workshop on Pre-Training: {{Perspectives}}, Pitfalls, and Paths Forward at {{ICML}} 2022}, 2022.

\bibitem[Schwarzer et~al.(2021{\natexlab{a}})Schwarzer, Anand, Goel, Hjelm, Courville, and Bachman]{schwarzerDataEfficientReinforcementLearning2021}
Max Schwarzer, Ankesh Anand, Rishab Goel, R.~Devon Hjelm, Aaron~C. Courville, and Philip Bachman.
\newblock Data-{{Efficient Reinforcement Learning}} with {{Self-Predictive Representations}}.
\newblock In \emph{9th {{International Conference}} on {{Learning Representations}}, {{ICLR}} 2021, {{Virtual Event}}, {{Austria}}, {{May}} 3-7, 2021}. {OpenReview.net}, 2021{\natexlab{a}}.

\bibitem[Schwarzer et~al.(2021{\natexlab{b}})Schwarzer, Rajkumar, Noukhovitch, Anand, Charlin, Hjelm, Bachman, and Courville]{schwarzerPretrainingRepresentationsDataEfficient2021}
Max Schwarzer, Nitarshan Rajkumar, Michael Noukhovitch, Ankesh Anand, Laurent Charlin, R.~Devon Hjelm, Philip Bachman, and Aaron~C. Courville.
\newblock Pretraining {{Representations}} for {{Data-Efficient Reinforcement Learning}}.
\newblock In Marc'Aurelio Ranzato, Alina Beygelzimer, Yann~N. Dauphin, Percy Liang, and Jennifer~Wortman Vaughan, editors, \emph{Advances in {{Neural Information Processing Systems}} 34: {{Annual Conference}} on {{Neural Information Processing Systems}} 2021, {{NeurIPS}} 2021, {{December}} 6-14, 2021, Virtual}, pages 12686--12699, 2021{\natexlab{b}}.

\bibitem[Seo et~al.(2022)Seo, Lee, James, and Abbeel]{seoReinforcementLearningActionfree2022}
Younggyo Seo, Kimin Lee, Stephen~L. James, and Pieter Abbeel.
\newblock Reinforcement learning with action-free pre-training from videos.
\newblock In Kamalika Chaudhuri, Stefanie Jegelka, Le~Song, Csaba Szepesv{\'a}ri, Gang Niu, and Sivan Sabato, editors, \emph{International Conference on Machine Learning, {{ICML}} 2022, 17-23 July 2022, Baltimore, Maryland, {{USA}}}, volume 162 of \emph{Proceedings of Machine Learning Research}, pages 19561--19579. {PMLR}, 2022.

\bibitem[Shafiullah and Pinto(2022)]{shafiullahOneAnotherLearning2022}
Nur~Muhammad Shafiullah and Lerrel Pinto.
\newblock One {{After Another}}: {{Learning Incremental Skills}} for a {{Changing World}}, March 2022.

\bibitem[Sharma et~al.(2020)Sharma, Gu, Levine, Kumar, and Hausman]{sharmaDynamicsawareUnsupervisedDiscovery2020}
Archit Sharma, Shixiang Gu, Sergey Levine, Vikash Kumar, and Karol Hausman.
\newblock Dynamics-aware unsupervised discovery of skills.
\newblock In \emph{8th International Conference on Learning Representations, {{ICLR}} 2020, Addis Ababa, Ethiopia, April 26-30, 2020}. {OpenReview.net}, 2020.

\bibitem[Silver et~al.(2017)Silver, Hubert, Schrittwieser, Antonoglou, Lai, Guez, Lanctot, Sifre, Kumaran, Graepel, Lillicrap, Simonyan, and Hassabis]{silverMasteringChessShogi2017}
David Silver, Thomas Hubert, Julian Schrittwieser, Ioannis Antonoglou, Matthew Lai, Arthur Guez, Marc Lanctot, Laurent Sifre, Dharshan Kumaran, Thore Graepel, Timothy Lillicrap, Karen Simonyan, and Demis Hassabis.
\newblock Mastering {{Chess}} and {{Shogi}} by {{Self-Play}} with a {{General Reinforcement Learning Algorithm}}.
\newblock \emph{arXiv:1712.01815 [cs]}, December 2017.

\bibitem[Srinivas and Abbeel(2021)]{srinivasUnsupervisedLearningReinforcement2021}
Aravind Srinivas and Pieter Abbeel.
\newblock Unsupervised {{Learning}} for {{Reinforcement Learning}}, 2021.

\bibitem[Stooke et~al.(2021)Stooke, Lee, Abbeel, and Laskin]{stookeDecouplingRepresentationLearning2021}
Adam Stooke, Kimin Lee, Pieter Abbeel, and Michael Laskin.
\newblock Decoupling {{Representation Learning}} from {{Reinforcement Learning}}, May 2021.

\bibitem[Sutton and Barto(1998)]{suttonReinforcementLearningIntroduction1998}
Richard~S. Sutton and Andrew~G. Barto.
\newblock Reinforcement {{Learning}}: {{An Introduction}}.
\newblock \emph{IEEE Trans. Neural Networks}, 9\penalty0 (5):\penalty0 1054--1054, 1998.
\newblock \doi{10.1109/TNN.1998.712192}.

\bibitem[Tassa et~al.(2018)Tassa, Doron, Muldal, Erez, Li, Casas, Budden, Abdolmaleki, Merel, Lefrancq, Lillicrap, and Riedmiller]{tassaDeepMindControlSuite2018}
Yuval Tassa, Yotam Doron, Alistair Muldal, Tom Erez, Yazhe Li, Diego de~Las Casas, David Budden, Abbas Abdolmaleki, Josh Merel, Andrew Lefrancq, Timothy Lillicrap, and Martin Riedmiller.
\newblock {{DeepMind Control Suite}}.
\newblock \emph{arXiv:1801.00690 [cs]}, January 2018.

\bibitem[Tassa et~al.(2020)Tassa, Tunyasuvunakool, Muldal, Doron, Trochim, Liu, Bohez, Merel, Erez, Lillicrap, and Heess]{tassaDmControlSoftware2020}
Yuval Tassa, Saran Tunyasuvunakool, Alistair Muldal, Yotam Doron, Piotr Trochim, Siqi Liu, Steven Bohez, Josh Merel, Tom Erez, Timothy Lillicrap, and Nicolas Heess.
\newblock Dm\_control: {{Software}} and {{Tasks}} for {{Continuous Control}}.
\newblock \emph{Software Impacts}, 6:\penalty0 100022, November 2020.
\newblock ISSN 26659638.
\newblock \doi{10.1016/j.simpa.2020.100022}.

\bibitem[van~den Oord et~al.(2019)van~den Oord, Li, and Vinyals]{oordRepresentationLearningContrastive2019}
Aaron van~den Oord, Yazhe Li, and Oriol Vinyals.
\newblock Representation {{Learning}} with {{Contrastive Predictive Coding}}, January 2019.

\bibitem[Xu et~al.(2022)Xu, {Parker-Holder}, Pacchiano, Ball, Rybkin, Roberts, Rockt{\"a}schel, and Grefenstette]{xuLearningGeneralWorld2022}
Yingchen Xu, Jack {Parker-Holder}, Aldo Pacchiano, Philip~J. Ball, Oleh Rybkin, Stephen~J. Roberts, Tim Rockt{\"a}schel, and Edward Grefenstette.
\newblock Learning {{General World Models}} in a {{Handful}} of {{Reward-Free Deployments}}, October 2022.

\bibitem[Yarats et~al.(2021)Yarats, Fergus, Lazaric, and Pinto]{yaratsReinforcementLearningPrototypical2021}
Denis Yarats, Rob Fergus, Alessandro Lazaric, and Lerrel Pinto.
\newblock Reinforcement {{Learning}} with {{Prototypical Representations}}.
\newblock In Marina Meila and Tong Zhang, editors, \emph{Proceedings of the 38th {{International Conference}} on {{Machine Learning}}, {{ICML}} 2021, 18-24 {{July}} 2021, {{Virtual Event}}}, volume 139 of \emph{Proceedings of {{Machine Learning Research}}}, pages 11920--11931. {PMLR}, 2021.

\bibitem[Yarats et~al.(2022)Yarats, Brandfonbrener, Liu, Laskin, Abbeel, Lazaric, and Pinto]{yaratsDonChangeAlgorithm2022}
Denis Yarats, David Brandfonbrener, Hao Liu, Michael Laskin, Pieter Abbeel, Alessandro Lazaric, and Lerrel Pinto.
\newblock Don't {{Change}} the {{Algorithm}}, {{Change}} the {{Data}}: {{Exploratory Data}} for {{Offline Reinforcement Learning}}.
\newblock In \emph{{{ICLR}} 2022 {{Workshop}} on {{Generalizable Policy Learning}} in {{Physical World}}}, 2022.

\bibitem[Yuan et~al.(2022)Yuan, Hao, Ni, Mu, Zheng, Hu, Liu, Chen, and Fan]{yuanEUCLIDEfficientUnsupervised2022}
Yifu Yuan, Jianye Hao, Fei Ni, Yao Mu, Yan Zheng, Yujing Hu, Jinyi Liu, Yingfeng Chen, and Changjie Fan.
\newblock {{EUCLID}}: {{Towards}} efficient unsupervised reinforcement learning with multi-choice dynamics model.
\newblock \emph{CoRR}, abs/2210.00498, 2022.
\newblock \doi{10.48550/arXiv.2210.00498}.

\bibitem[Zeng et~al.(2022)Zeng, Zhang, Chen, Liang, and Yang]{zengAPDLearningDiverse2022}
Kailin Zeng, QiYuan Zhang, Bin Chen, Bin Liang, and Jun Yang.
\newblock {{APD}}: {{Learning Diverse Behaviors}} for {{Reinforcement Learning Through Unsupervised Active Pre-Training}}.
\newblock \emph{IEEE Robotics and Automation Letters}, 7\penalty0 (4):\penalty0 12251--12258, October 2022.
\newblock ISSN 2377-3766, 2377-3774.
\newblock \doi{10.1109/LRA.2022.3214057}.

\bibitem[Zhang et~al.(2020)Zhang, Cai, Huang, and Li]{zhangExplorationMaximizingEnyi2020}
Chuheng Zhang, Yuanying Cai, Longbo Huang, and Jian Li.
\newblock Exploration by {{Maximizing R}}\textbackslash 'enyi {{Entropy}} for {{Reward-Free RL Framework}}, December 2020.

\bibitem[Zhao et~al.(2022)Zhao, Lin, Li, Liu, and Huang]{zhaoMixtureSurprisesUnsupervised2022}
Andrew Zhao, Matthieu~Gaetan Lin, Yangguang Li, Yong-Jin Liu, and Gao Huang.
\newblock A mixture of surprises for unsupervised reinforcement learning.
\newblock In \emph{Advances in {{Neural Information Processing Systems}} 35: {{Annual Conference}} on {{Neural Information Processing Systems}} 2022, {{NeurIPS}} 2022}, December 2022.

\end{thebibliography}

\clearpage

\appendix

\section{Extended Demonstration on PointMass}
\label{sec:extended_demo_pointmass}

\begin{figure*}[ht!]
    \centering
    \begin{subfigure}{0.5\textwidth}
        \includegraphics[width=\textwidth]{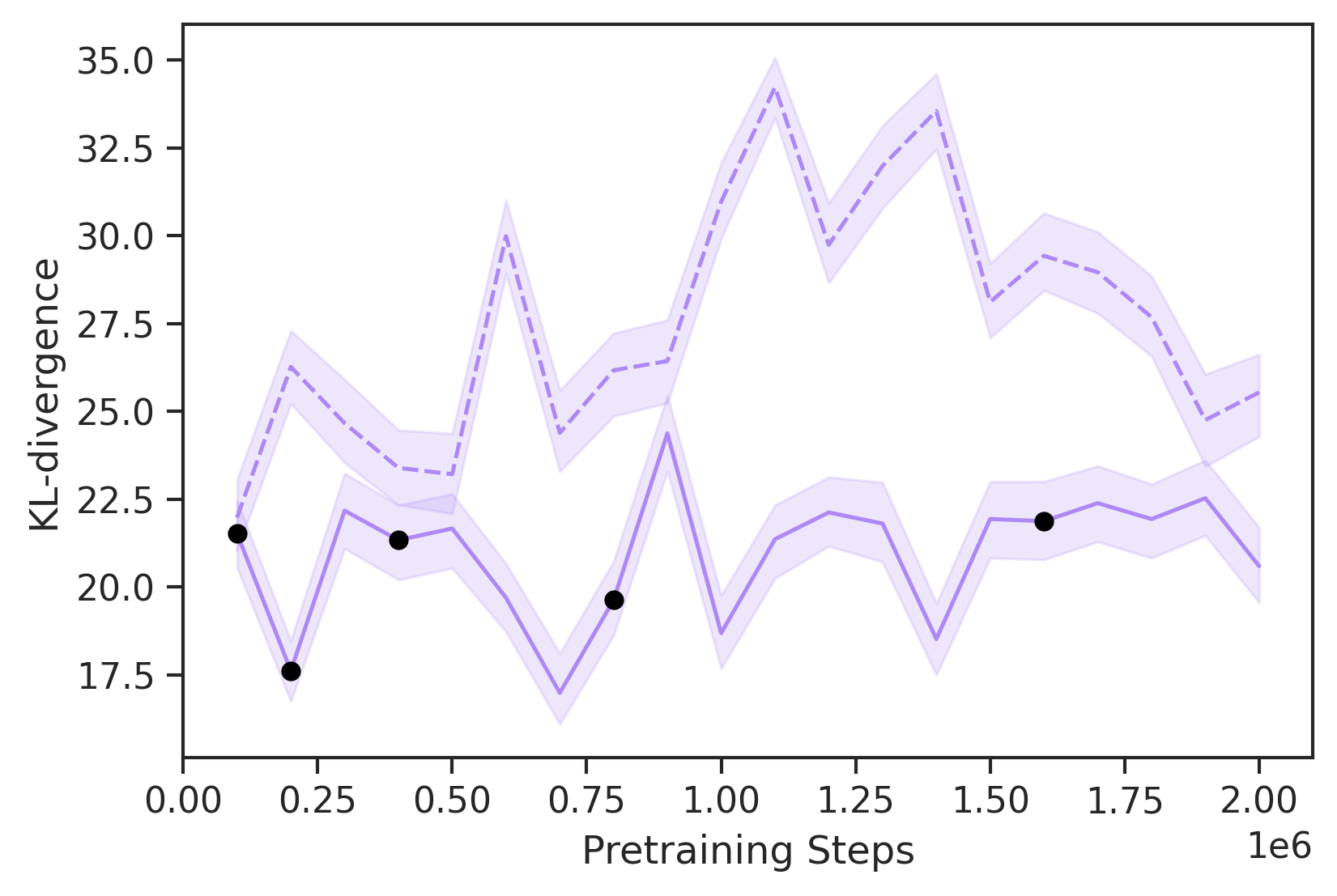}
        \caption{Pretraining ($\SI{2}{\mega\nothing}$ steps)}
        \label{fig:pointmass_pretraining}
    \end{subfigure}~
    \begin{subfigure}{0.45\textwidth}
        \includegraphics[width=\textwidth]{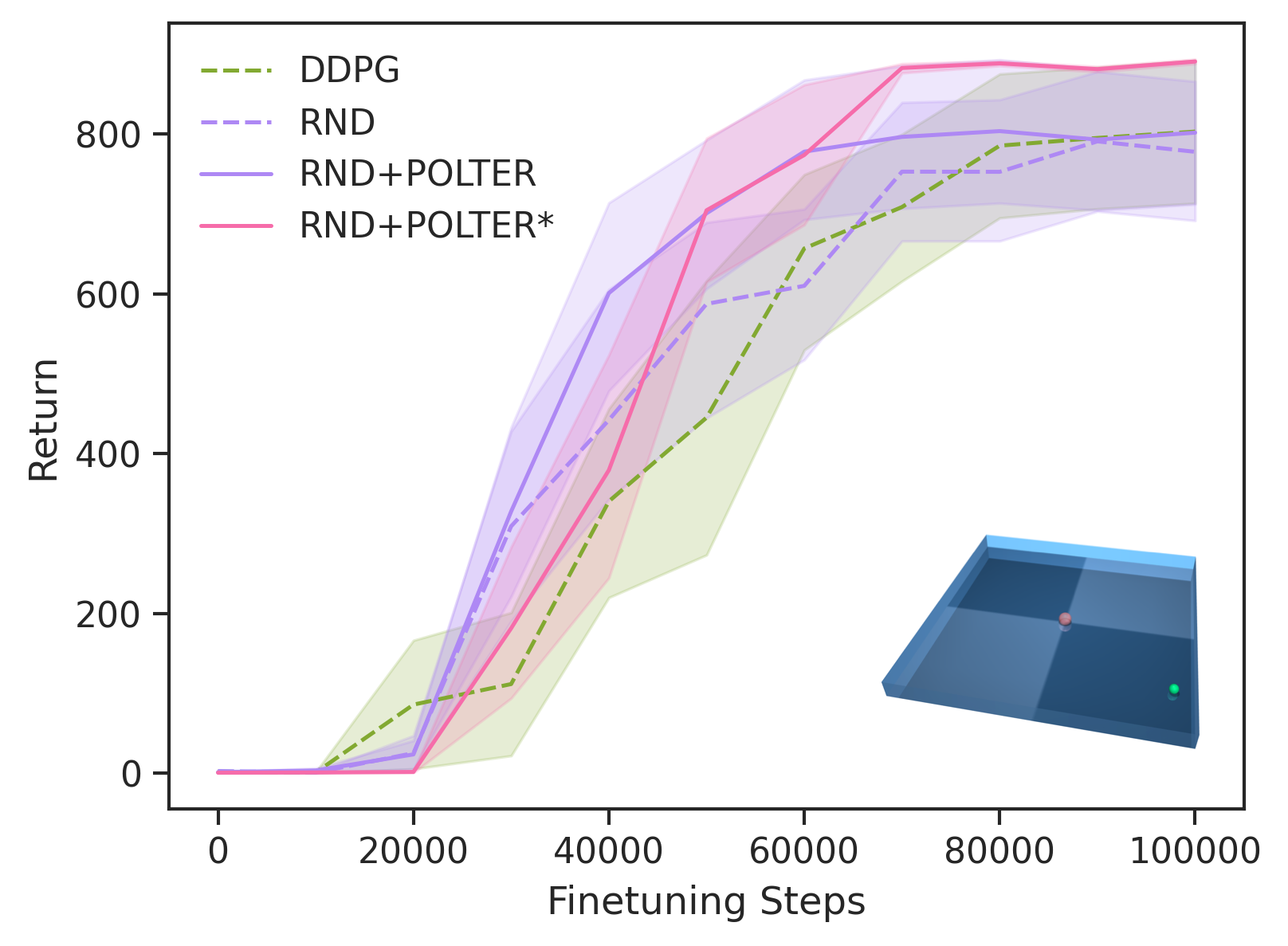}
        \caption{Finetuning ($\SI{100}{\kilo\nothing}$ steps)}
        \label{fig:pointmass_finetuning}
    \end{subfigure}\\
    ~
    \begin{subfigure}{0.5\textwidth}
        \includegraphics[width=\textwidth]{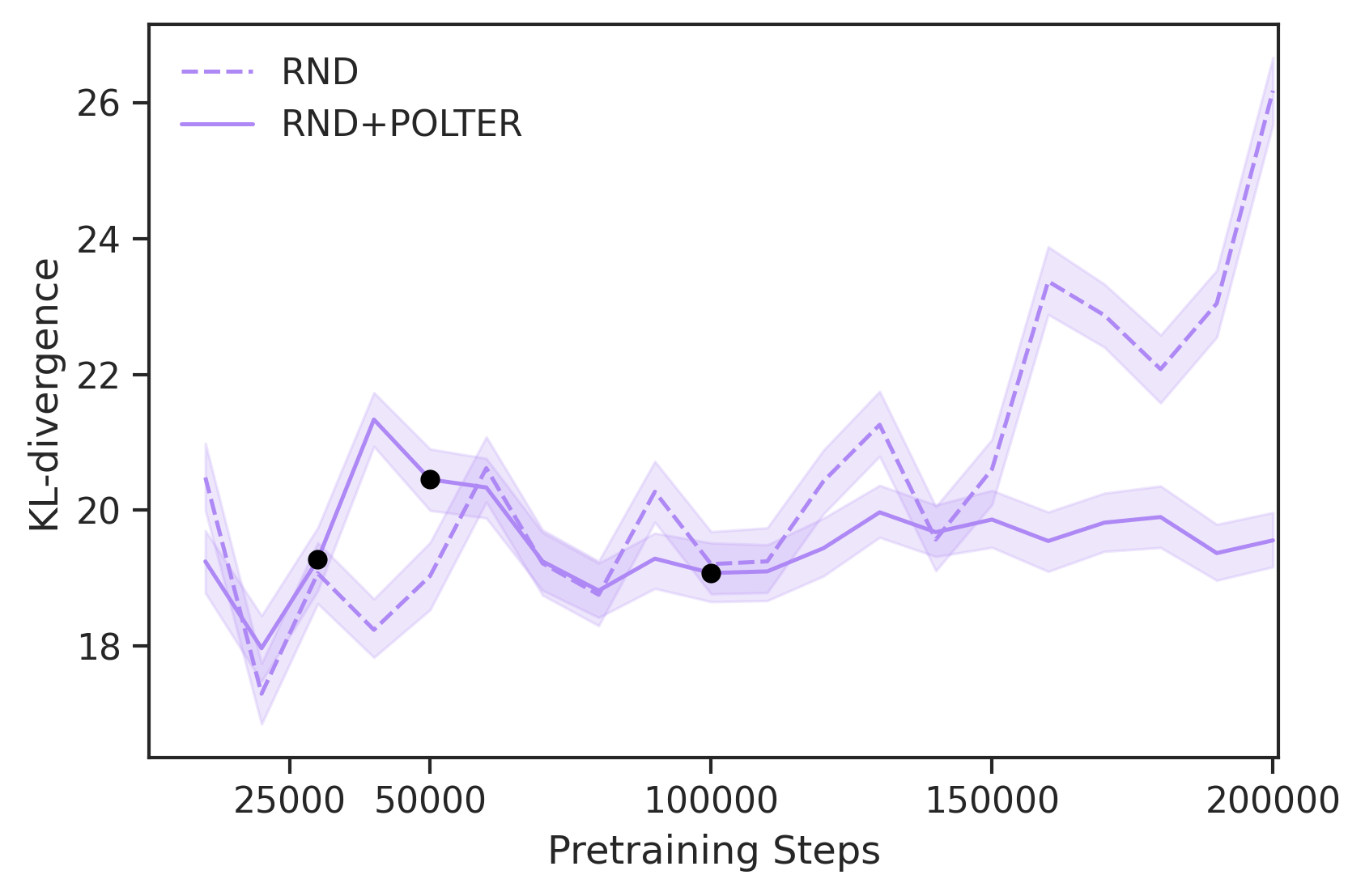}
        \caption{Pretraining  ($\SI{200}{\kilo\nothing}$ steps)}
        \label{fig:pointmass_pretraining_200k}
    \end{subfigure}~
    \begin{subfigure}{0.45\textwidth}
        \includegraphics[width=\textwidth]{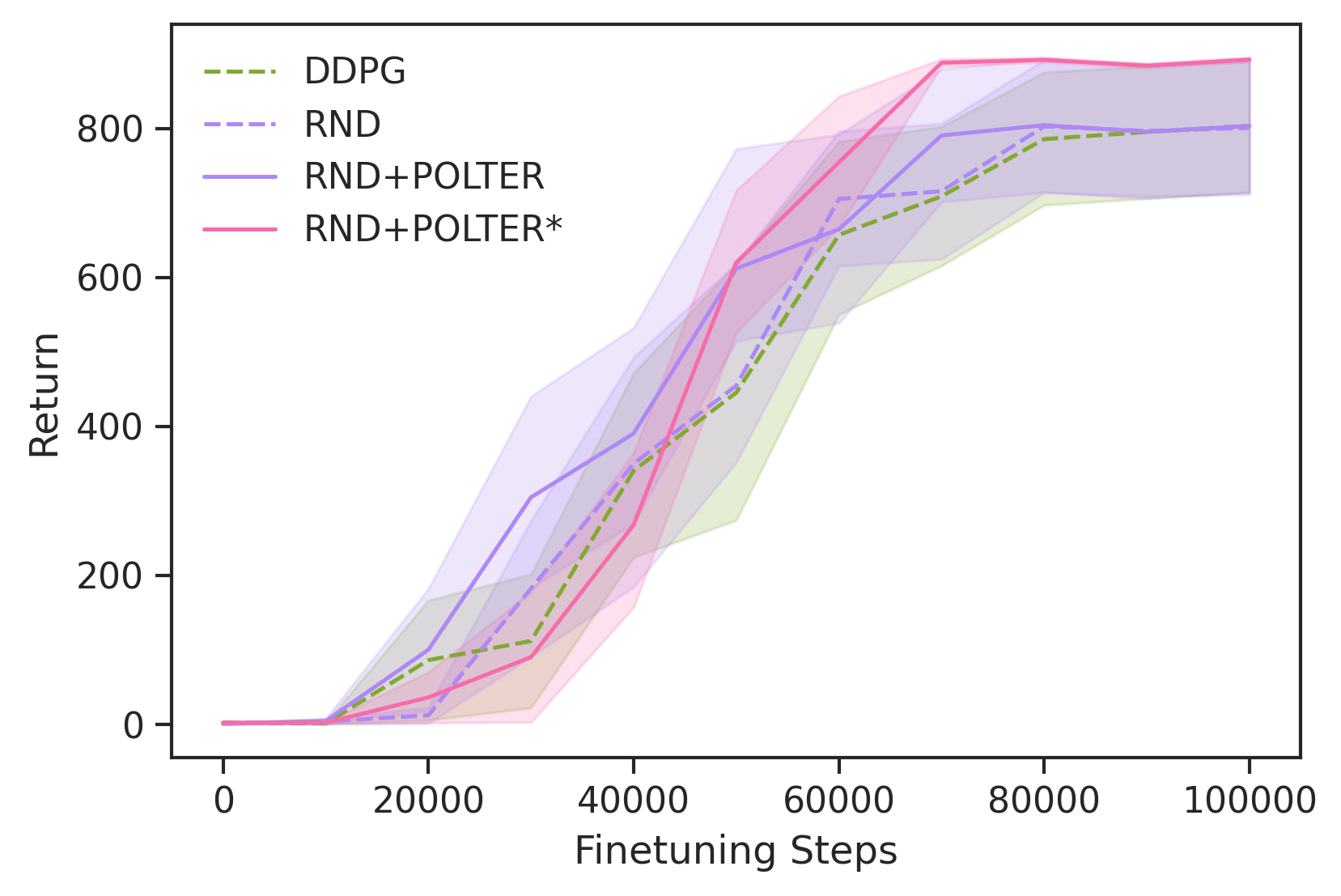}
        \caption{Finetuning  ($\SI{100}{\kilo\nothing}$ steps)}
        \label{fig:pointmass_finetuning_200k}
    \end{subfigure}
    \caption{\textbf{(\subref{fig:pointmass_pretraining})},  \textbf{(\subref{fig:pointmass_pretraining_200k})} Average \ac{KL-divergence} of \ac{RND} (dashed) and \ac{RND}+\polter (solid) between the pretraining policy $\pi(s_0)$ and the optimal pretraining policy $\pi^*_T(s_0)$ in the PointMass domain during reward-free pretraining. \textbf{(\subref{fig:pointmass_pretraining})} is trained for $\SI{2}{\mega\nothing}$ steps, and \textbf{(\subref{fig:pointmass_pretraining_200k})} is trained for $\SI{200}{\kilo\nothing}$. Each policy is evaluated every $\SI{100}{\kilo\nothing}$ steps on \num{20} initial states over \num{10} seeds. The black dots indicate the steps a snapshot is added to the ensemble. \textbf{(\subref{fig:pointmass_finetuning})}, \textbf{(\subref{fig:pointmass_finetuning_200k})} Return during finetuning after $\SI{2}{\mega\nothing}$ and $\SI{200}{\kilo\nothing}$ pretraining steps, where the target is placed at a fixed random position for each of the 10 seeds. We also provide a \ac{DDPG} baseline without pretraining and \ac{RND}+POLTER* using the optimal policy instead of the ensemble. The shaded area indicates the standard error.}
\end{figure*}

We evaluate the effect of our regularization on \ac{RND}~\citep{burdaExplorationRandomNetwork2019}, a well-known knowledge-based \ac{URL} algorithm, in the simplistic PointMass environment~\citep{tassaDeepMindControlSuite2018,tassaDmControlSoftware2020}.
In this environment, the agent has to apply a force to a mass (green in \cref{fig:pointmass_finetuning}) in a 2D plane to reach a target (red) and observes its position and speed.
We pretrain a \ac{DDPG} agent using \ac{RND} with and without \polter for $\SI{2}{\mega\nothing}$ steps and evaluate the policies every $\SI{100}{\kilo\nothing}$ steps, in accordance with the URL benchmark evaluation protocol.
Because PointMass is a much simpler environment than the ones contained in URLB and to gain a better resolution for the start of the pretraining, we repeat the experiment with $\SI{200}{\kilo\nothing}$ pretraining and $\SI{100}{\kilo\nothing}$ finetuning steps.
We use \num{10} seeds.

In our whitebox benchmark PointMass, we can train the optimal prior policy and compute the \ac{KL-divergence} between the optimal prior policy and the current pretraining policy $\infdiv{\pi^*_T}{\pi}$.
Comparing the \ac{KL-divergence} during pretraining with and without \polter in \cref{fig:pointmass_pretraining} shows that with our method, the \ac{KL-divergence} to the optimal prior policy $\pi^*_T$ is generally lower, which is also apparent for the short pretraining (\cref{fig:pointmass_pretraining_200k}).
POLTER keeps the KL-divergence at bay, whereas RND without POLTER diverges much more, which is consistent with prior work showing that URL tends to diverge with too many pretraining steps.

\subsection{Effect of \polter on Finetuning}
The effect of this improved prior is also apparent during finetuning in \cref{fig:pointmass_finetuning}.
Here, we see that using an ensemble is a good proxy for the optimal prior due to the similar sample-efficiency.
In addition, we observe a speed-up of approximately  $\SI{25}{\percent}$ ($\SI{2}{\mega\nothing}$ pretraining steps) and $\SI{12.5}{\percent}$ ($\SI{200}{\kilo\nothing}$ pretraining steps)  when using RND+POLTER in reaching the final performance of DDPG.

\subsection{Visualization of State Distributions}
\begin{figure*}[ht!]
    \centering
     \begin{subfigure}[b]{0.68\textwidth}
         \centering
         \includegraphics[width=\linewidth]{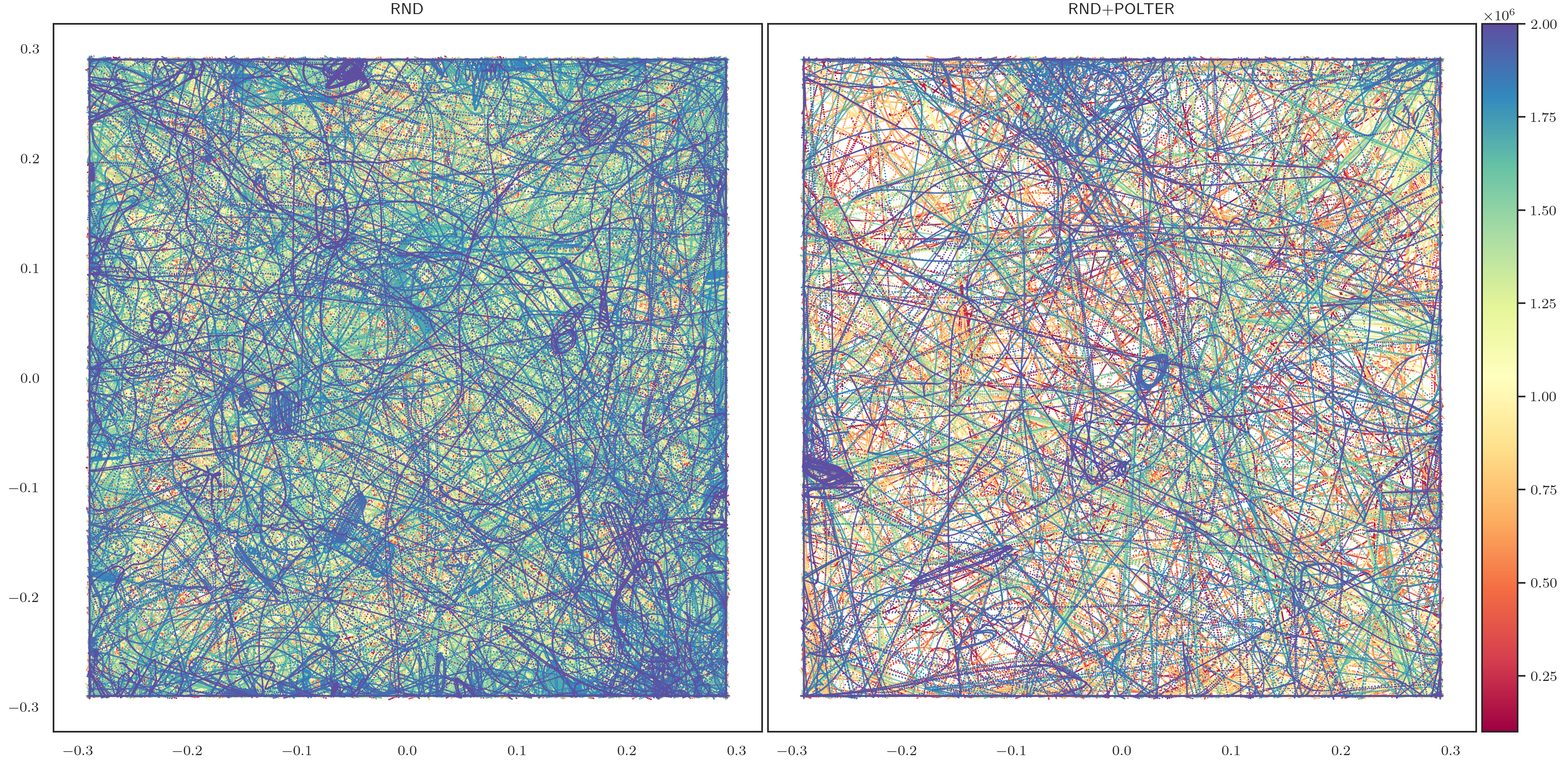}
        \caption{State distribution $\rho(s)$ of several checkpoints.}
        \label{fig:pointmass_state_distribution}
     \end{subfigure}
     \hfill
     \centering
     \begin{subfigure}[b]{0.68\textwidth}
         \centering
         \includegraphics[width=\linewidth]{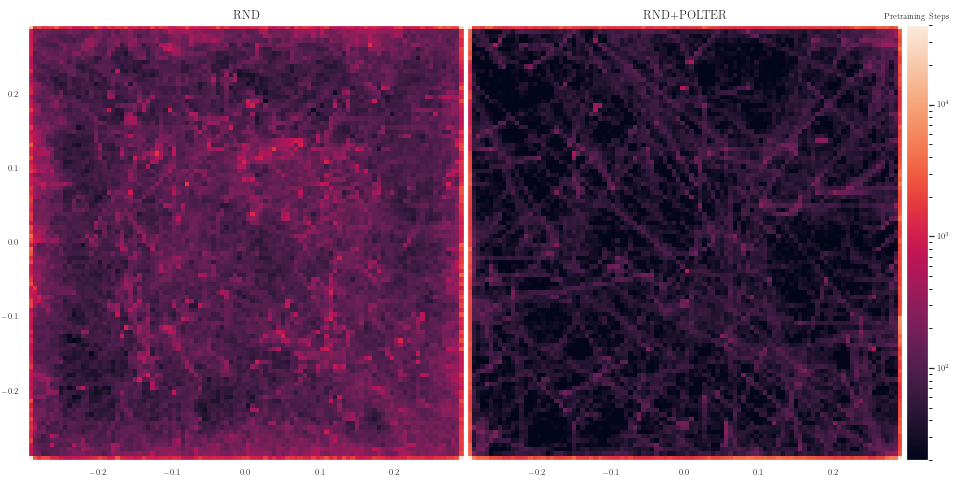}
        \caption{Discretized state histogram of the positions. Looking closely, we can see that the edges of the 2D plane are very often visited. This is due to policies constantly applying the same force and thus reaching the edge of the 2D plane.}
        \label{fig:pointmass_position}
     \end{subfigure}
     \hfill
     \centering
     \begin{subfigure}[b]{0.68\textwidth}
         \centering
         \includegraphics[width=\linewidth]{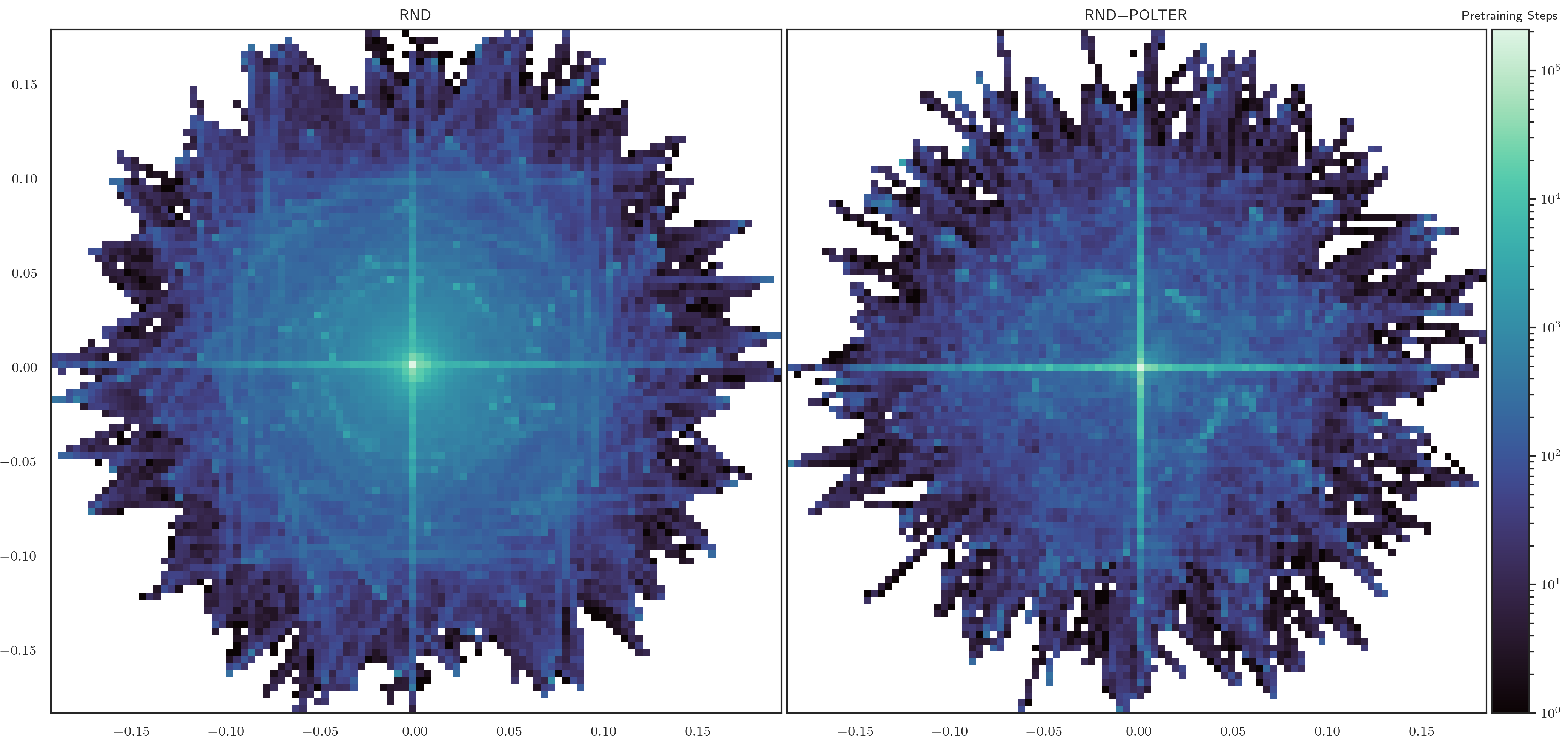}
        \caption{Discretized state histogram of the speeds. Note that the prominent horizontal and vertical lines result from moving at the edges and being stuck at the corners of the 2D plane.}
        \label{fig:pointmass_speed}
     \end{subfigure}
    \caption{State distribution and histogram of RND (always left column) and RND+\polter (right column) during pretraining on the PointMass environment.}
    \label{fig:my_label}
\end{figure*}

In \cref{fig:pointmass_state_distribution}, we can see the state distribution changes during pretraining with and without \polter.
With \polter, the state space coverage is less, and the trajectories seem more ordered.
RND without \polter also seems to visit the edges often at the end.
When using the \polter regularization, we can see that each pretraining checkpoint is visiting different states, as indicated by the visibility of the previous checkpoint's state visitations.
When not using \polter, we can see that the visitations overlap.
\cref{fig:pointmass_position,fig:pointmass_speed} show the discretized position and speed the agent explores throughout pretraining.
Especially the discretized speed (\cref{fig:pointmass_speed}) demonstrates the tradeoff between dampening the exploration with \polter and finding better prior policies because with \polter, fewer states are frequented, and the states are less extreme. 

\section{Extended Experiments on Pendulum}
\label{sec:pendulum_demo}

\begin{figure*}[ht!]
    \centering
    \includegraphics[width=0.5\textwidth]{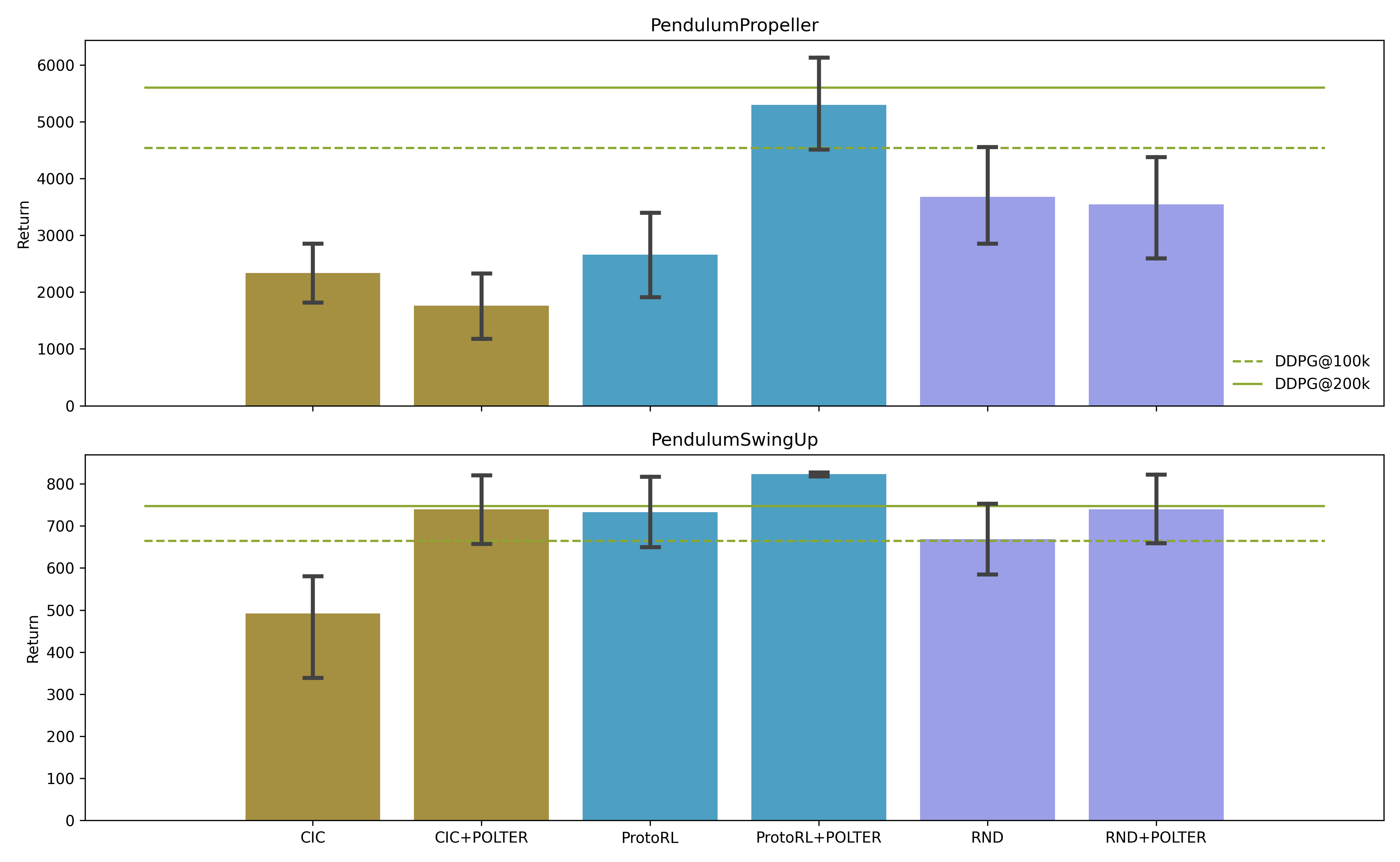}
    \caption{Average return over 10 seeds on two tasks in the Pendulum domain with different characteristics. Algorithms from all three \ac{URL} categories are compared to a \ac{DDPG} baseline that is trained for either $\SI{100}{\kilo\nothing}$ or $\SI{200}{\kilo\nothing}$ steps without pretraining.}
    \label{fig:pendulum_results}
\end{figure*}

In this experiment, we shed further light on the properties of \polter and its effect under different finetuning tasks. For example, in the PendulumPropeller task, the goal is to maximize the angular velocity of the pole. However, this requires a rather extreme policy, whereas, in the PendulumSwingUp task, the agent has to balance the pole and keep it upright.

The SwingUp task demonstrates that \polter can consistently improve the performance of the baseline \ac{URL} agent. However, for the Propeller task, the results are mixed. For \ac{ProtoRL}, \polter again shows a great improvement over the baseline. This supports the performance that we observed in the main experiments. But for \ac{CIC} and \ac{RND}, \polter decreases performance. The extreme policy of achieving a high angular velocity is further away from the average, so applying \polter has a detrimental effect. Nevertheless, note that regardless of this failure case, on average, the tasks will lie closer to the average state distribution, and thus, \polter will increase the performance.

\section{State-Visitation Entropy}
\label{sec:entropy_polter}

To gain further insights into \polter, we conduct an experiment following \citet{hazanProvablyEfficientMaximum2019}.
We discretize the state space of the Walker environment and compute the state-visitation entropy during reward-free pretraining.
In \cref{tab:entropy}, we see that the entropy of the distribution of \polter regularized algorithms is lower than that of their counterpart.
This effect is most pronounced in data-based algorithms, such as \ac{ProtoRL} and \ac{APT}, where the performance is also improved the most (see \cref{tab:agg_statistics}).

\begin{table*}[ht!]
\centering
\caption{State-visitation entropy of the evaluated \ac{URL} algorithm categories in the Walker domain during pretraining. Averaged over \num{10} seeds with $\SI{50}{\kilo\nothing}$ states each at pretraining steps $\SI{100}{\kilo\nothing}$, $\SI{500}{\kilo\nothing}$, $\SI{1}{\mega\nothing}$ and $\SI{2}{\mega\nothing}$.}
\begin{tabular}{cccc}
\toprule
\polter  &   Data &  Knowledge &  Competence \\
\midrule
\cross  & $0.2772 \pm 0.0188$ &     $0.2863 \pm 0.0036$ &      $0.2540 \pm 0.0429$  \\
\tick   & $0.2545 \pm 0.0493$ &     $0.2848 \pm 0.0054$ &      $0.2511 \pm 0.0422$  \\
\bottomrule
\end{tabular}
\label{tab:entropy}
\end{table*}

Knowledge-based algorithms also benefit from the regularization but have a slightly reduced entropy.
Because competence-based algorithms already average over a set of skills found during pretraining, the effect of \polter is the smallest.

These results imply that \polter does not lead to a better state-space exploration.
Instead, its performance gains are the result of an improved prior as indicated by the reduced \ac{KL-divergence} between the policy and the optimal pretraining policy on PointMass (\cref{sec:pointmass_demo}).
This experiment showed that a good exploration of the state-space is required but not sufficient to achieve good performance with \ac{URL} algorithms.

\section{Environments in the \acl{URLB}}

The \acl{URLB}~\cite{laskinURLBUnsupervisedReinforcement2021} contains three domains with the topics of locomotion and manipulation.
\textbf{Walker}, \textbf{Quadruped} and \textbf{Jaco}, are used to explore the effects of different \ac{URL} algorithms.
It has a specific training and evaluation protocol which we also follow in this work.
The \textbf{Walker} domain contains a planar walker constrained to a 2D vertical plane, with an 18-dimensional observation space and a 6-dimensional action space with three actuators for each leg.
The associated tasks are \textit{stand}, \textit{walk}, \textit{run} and \textit{flip}.
The walker domain provides a challenging start for the agent since it needs to learn balancing and locomotion skills to be able to adapt to the given tasks.
The next domain is \textbf{Quadruped}, which expands to the 3D space.
It has a much larger state space of 56 dimensions and a 12-dimensional action space with three actuators for each leg.
The tasks in this environment are \textit{stand}, \textit{walk}, \textit{jump} and \textit{run}.
The last environment used is the \textbf{Jaco Arm}, which is a robotic arm with 6-DOF and a three-finger gripper.
This domain is very different from the other two, as its setting is manipulation and not locomotion. 
The tasks are \textit{Reach top left}, \textit{Reach top right}, \textit{Reach bottom left} and \textit{Reach bottom right}.

\section{Hyperparameters and Resources}
\label{sec:hyperparameters}
\paragraph{\polter Hyperparameters}
\label{sec:polter_setup}

During pretraining, we construct the mixture ensemble policy $\tilde\pi$ with $k=7$ members at specific time steps $\mathcal{T}_E$.
For adding each member we choose the ensemble snapshot time steps $\mathcal{T}_E = \{\SI{25}{\kilo\nothing}, \SI{50}{\kilo\nothing}, \SI{100}{\kilo\nothing}, \SI{200}{\kilo\nothing}, \SI{400}{\kilo\nothing}, \SI{800}{\kilo\nothing}, \SI{1.6}{\mega\nothing}\}$.
The steps were chosen according to initial experiments of applying \ac{RND} in the Quadruped domain, where there are large changes of the intrinsic reward at the beginning, which become progressively smaller over time.
We set the regularization strength $\alpha = 1$ and use the same hyperparameters for each of the three domains unless specified otherwise.

\paragraph{Baseline Hyperparameters}
The hyperparameters for our baseline algorithms follow \citet{laskinURLBUnsupervisedReinforcement2021} and \citet{laskinCICContrastiveIntrinsic2022}.
The hyperparameters for the \ac{DDPG} baseline agent are described in \cref{tab:ddpg_hyperparameters}.

\begin{table}[ht!]
\centering
\caption{Hyperparameters for the \ac{DDPG} algorithm.}
\begin{tabular}{l r}
\toprule
Hyperparameter & Value \\
\midrule
Replay buffer capacity & $1 \times 10^6$ \\
Action repeat & $1$ \\
Seed frames & $4000$ \\
$n$-step returns & $3$ \\
Batch size & $1024$ \\
Discount factor $\gamma$ & $0.99$ \\
Optimizer & Adam \\
Learning rate & $1 \times 10^{-4}$ \\
Agent update frequency & $2$ \\
Critic target EMA rate & $0.01$ \\
Feature size & $1024$ \\
Hidden size & $1024$ \\
Exploration noise std clip & $0.3$ \\
Exploration noise std value & $0.2$ \\
Pretraining frames & $2 \times 10^6$ \\
Finetuning frames & $1 \times 10^5$ \\
\bottomrule
\end{tabular}
\label{tab:ddpg_hyperparameters}
\end{table}

\paragraph{Compute Resources}
\label{sec:compute_resources}

All experiments were run on our internal compute cluster on NVIDIA RTX 1080 Ti and NVIDIA RTX 2080 Ti GPUs and had 64GB of RAM and 10 CPU cores.
In total, we trained over \num{12000} models and performed $\approx$ \num{3500200000} environment steps.

\section{Detailed Results on \acl{URLB}}
\label{sec:detailed_results_urlb}

\begin{figure*}[ht!]
    \centering
    \includegraphics[width=0.7\textwidth]{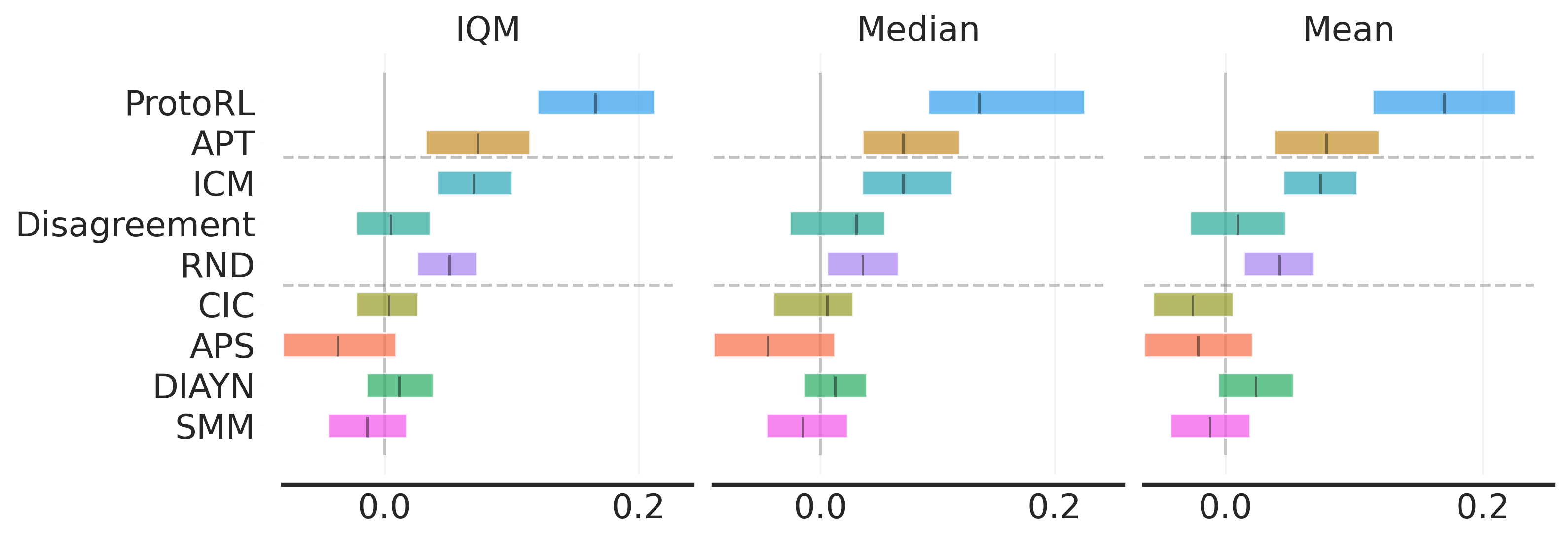}
    \caption{Aggregate statistics of the absolute improvement  with \polter per \ac{URL} category.}
    \label{fig:relative_improvement_iqm_categories_perpoint}
\end{figure*}

\begin{figure*}[ht!]
     \centering
     \hspace*{\fill}\begin{subfigure}[b]{0.4\textwidth}
         \centering
         \includegraphics[width=\linewidth]{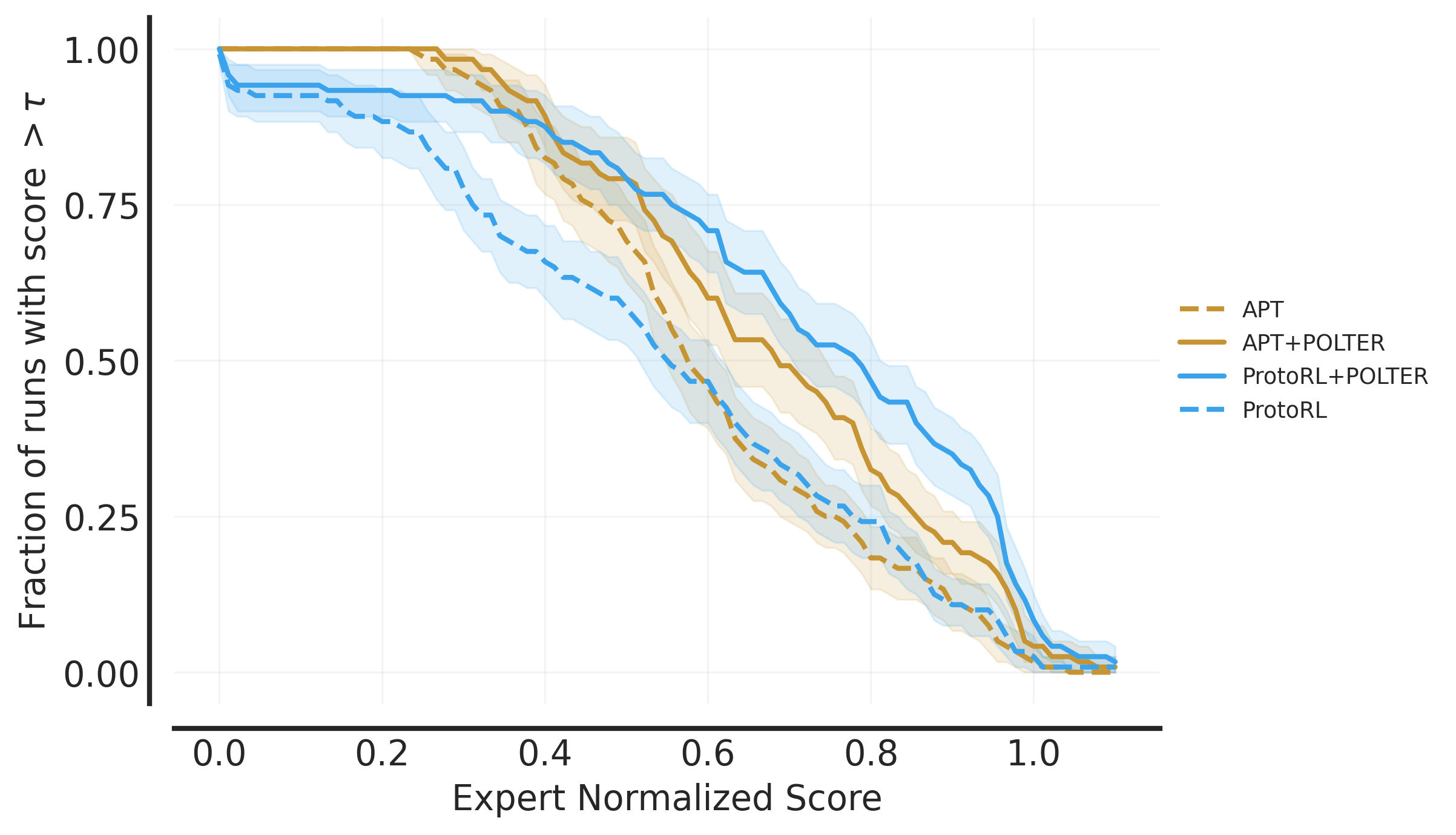}
        \label{fig:perf_profile_data}
     \end{subfigure}
     \hfill
     \begin{subfigure}[b]{0.4\textwidth}
         \centering
         \includegraphics[width=\linewidth]{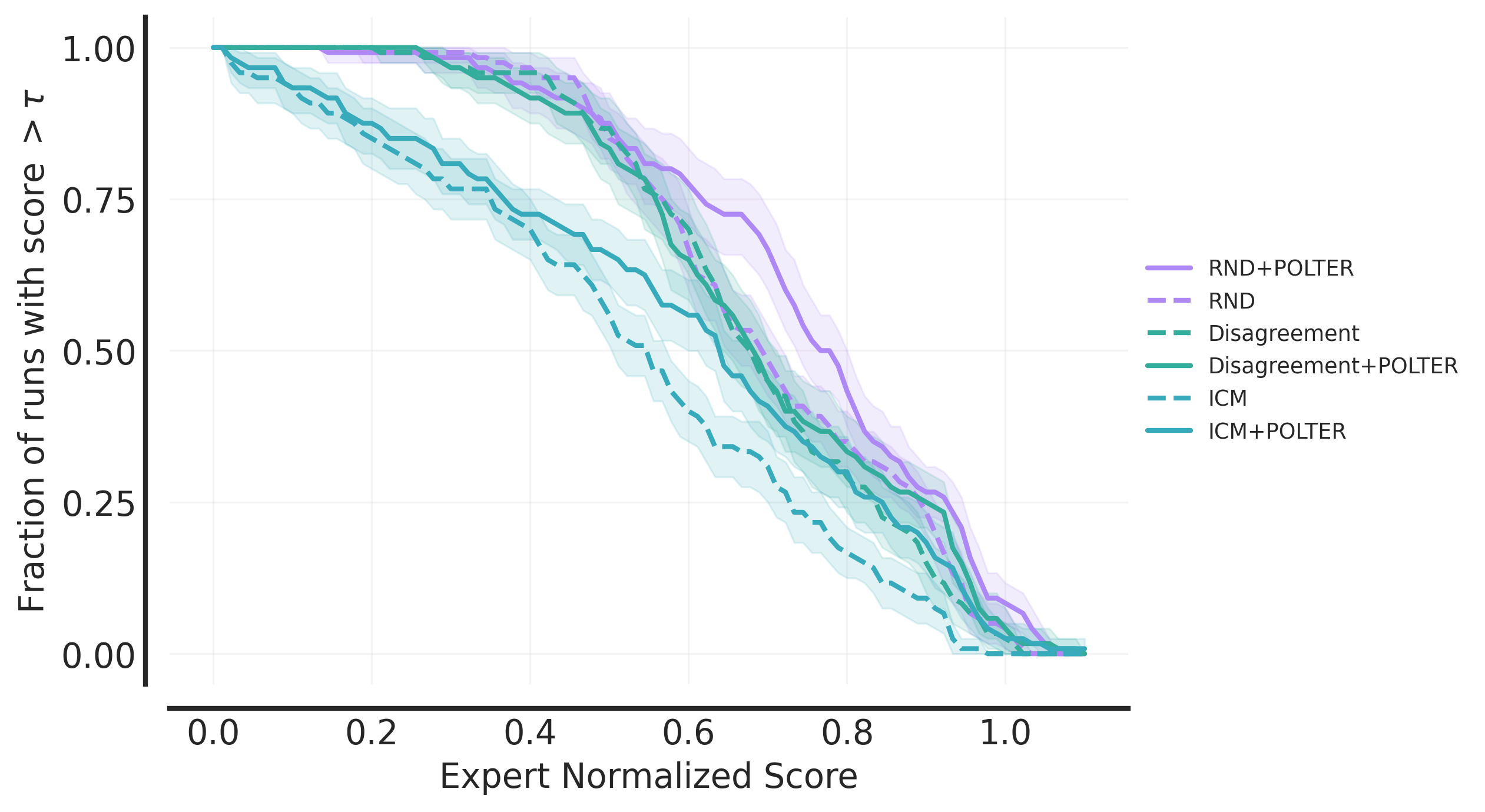}
        \label{fig:perf_profile_knowledge}
     \end{subfigure}
     \hspace*{\fill}\\
     \begin{subfigure}[b]{0.4\textwidth}
         \centering
         \includegraphics[width=\linewidth]{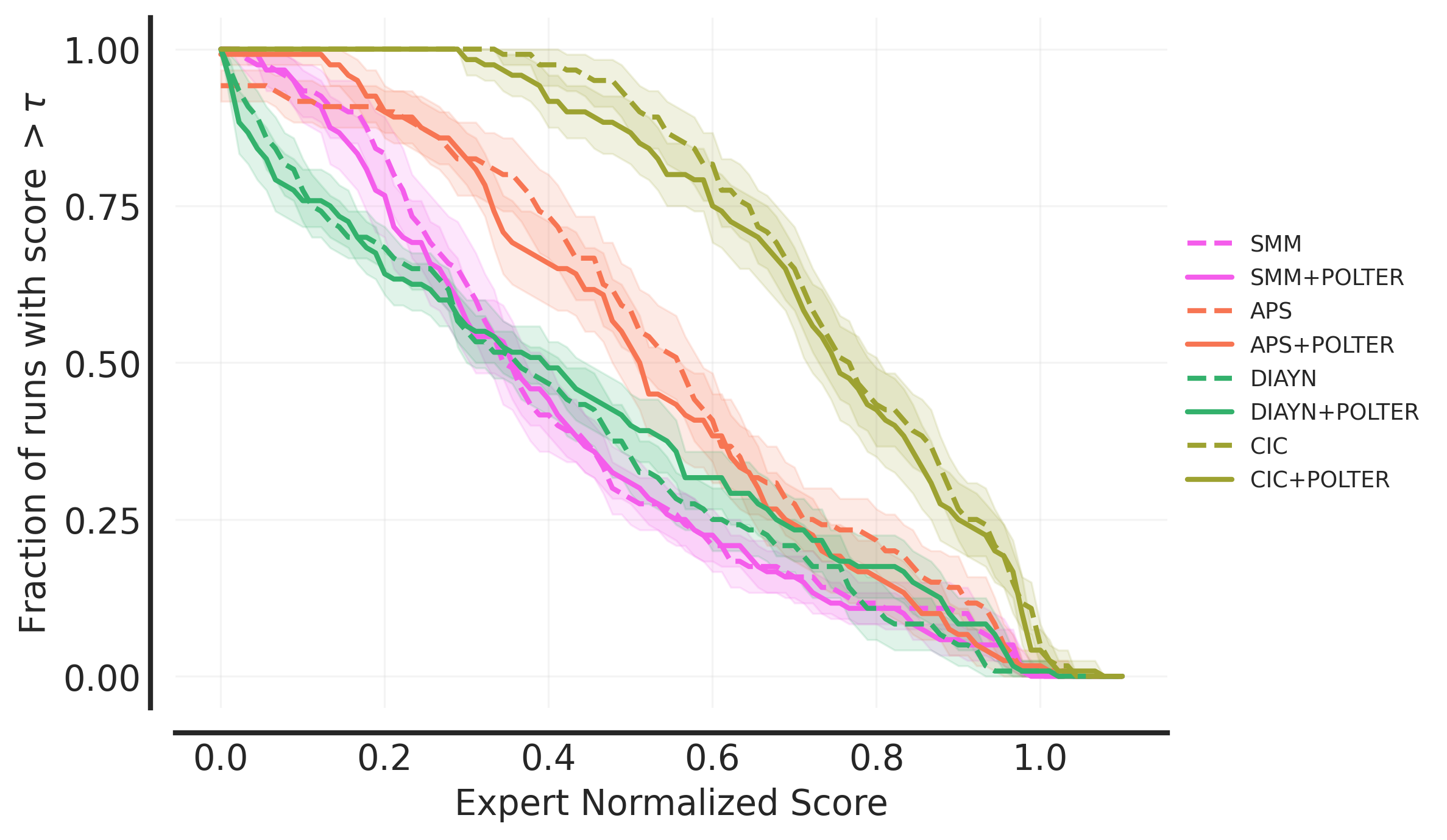}
        \label{fig:perf_profile_skill}
     \end{subfigure}
    \caption{Performance profiles after finetuning of the different algorithms averaged over 10 seeds where the shaded region indicates the standard error. Variants without \polter are dashed.}
    \label{fig:perf_profiles}
\end{figure*}

\begin{figure*}[ht!]
    \centering
    \includegraphics[width=0.7\textwidth]{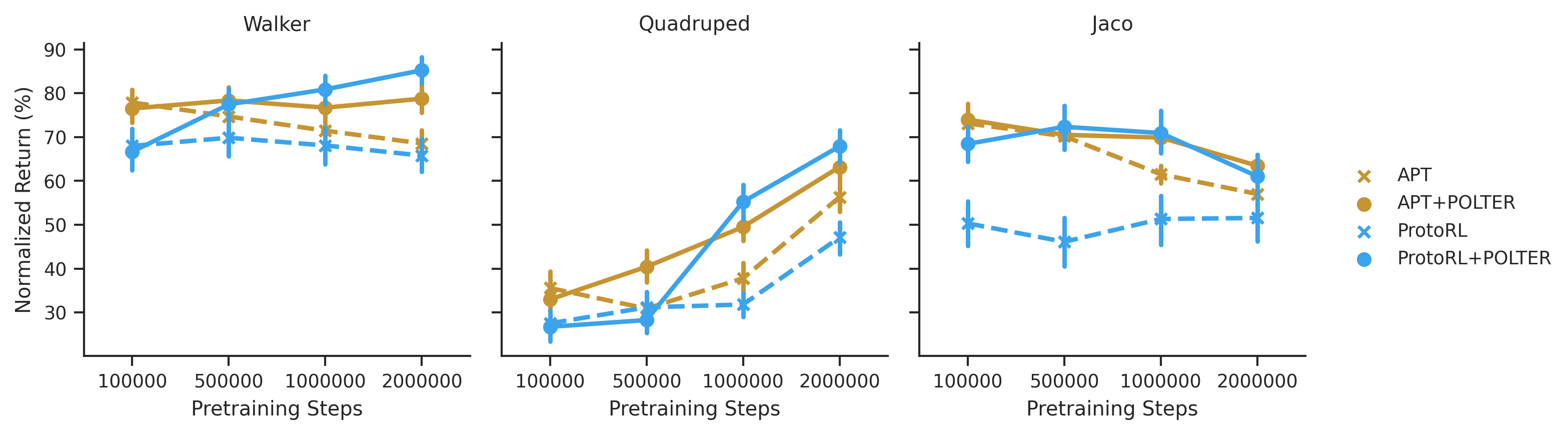}
    \includegraphics[width=0.7\textwidth]{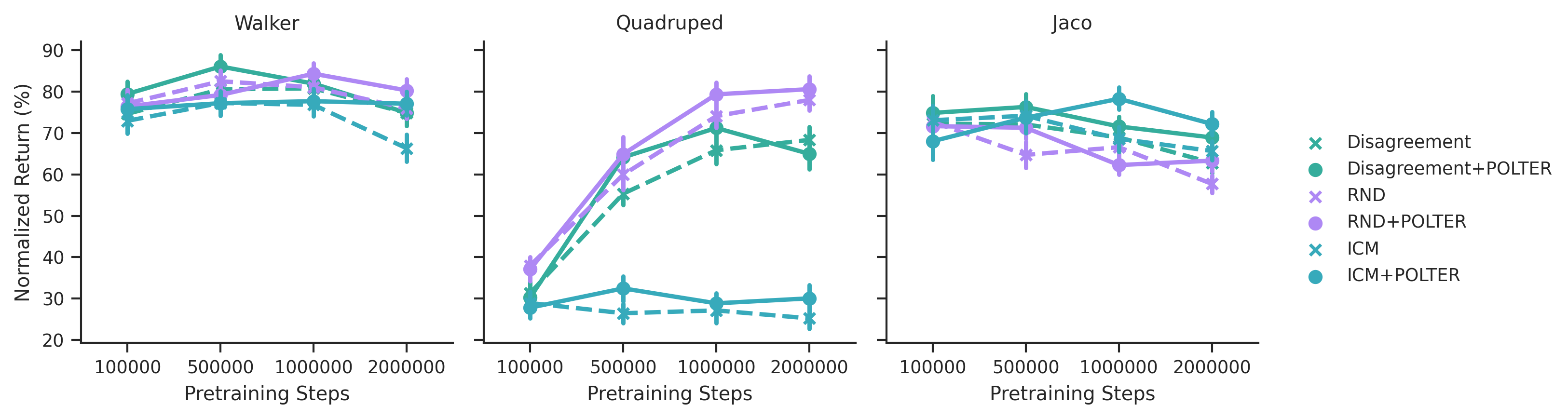}
    \includegraphics[width=0.7\textwidth]{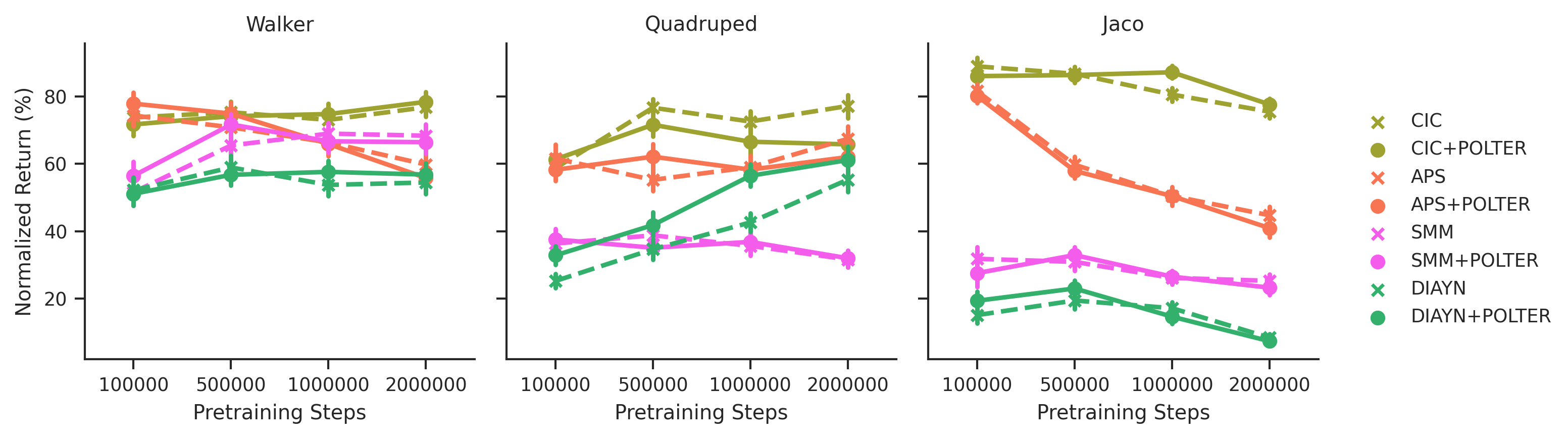}
    \caption{Finetuning from different pretraining snapshots of data-, knowledge- and competence-based algorithms. The error bars indicate the standard error of the mean.}
    \label{fig:pretraining_summary}
\end{figure*}

\begin{figure*}[ht!]
    \centering
    \includegraphics[width=0.8\textwidth]{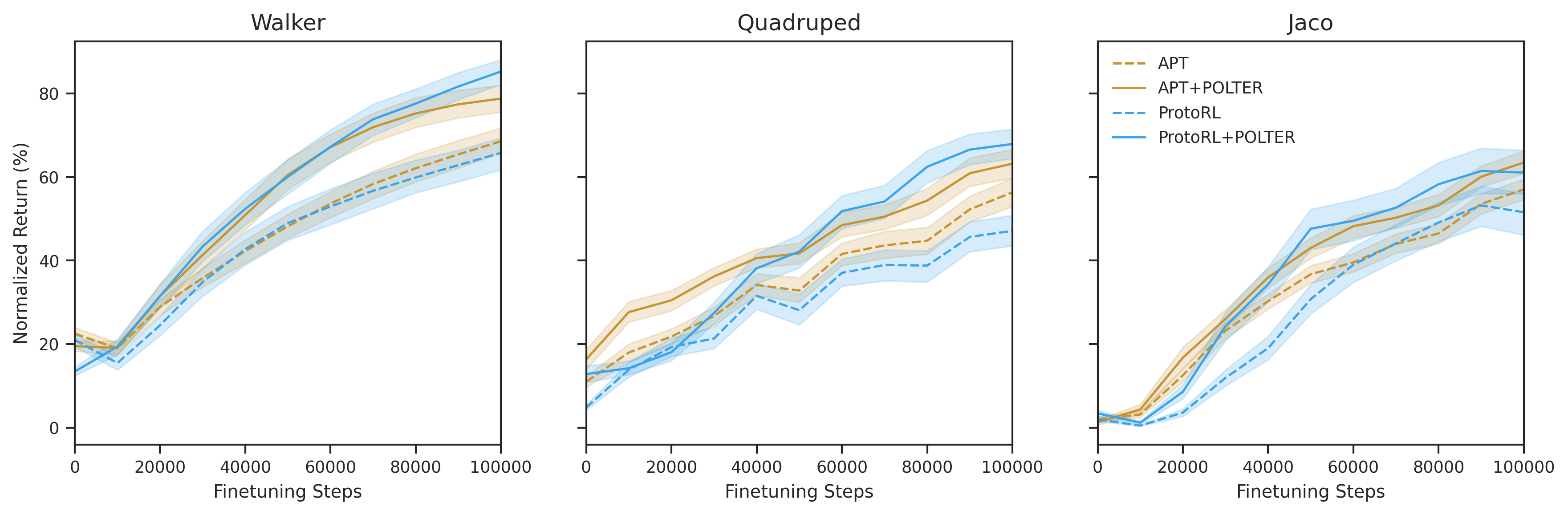}
    \includegraphics[width=0.8\textwidth]{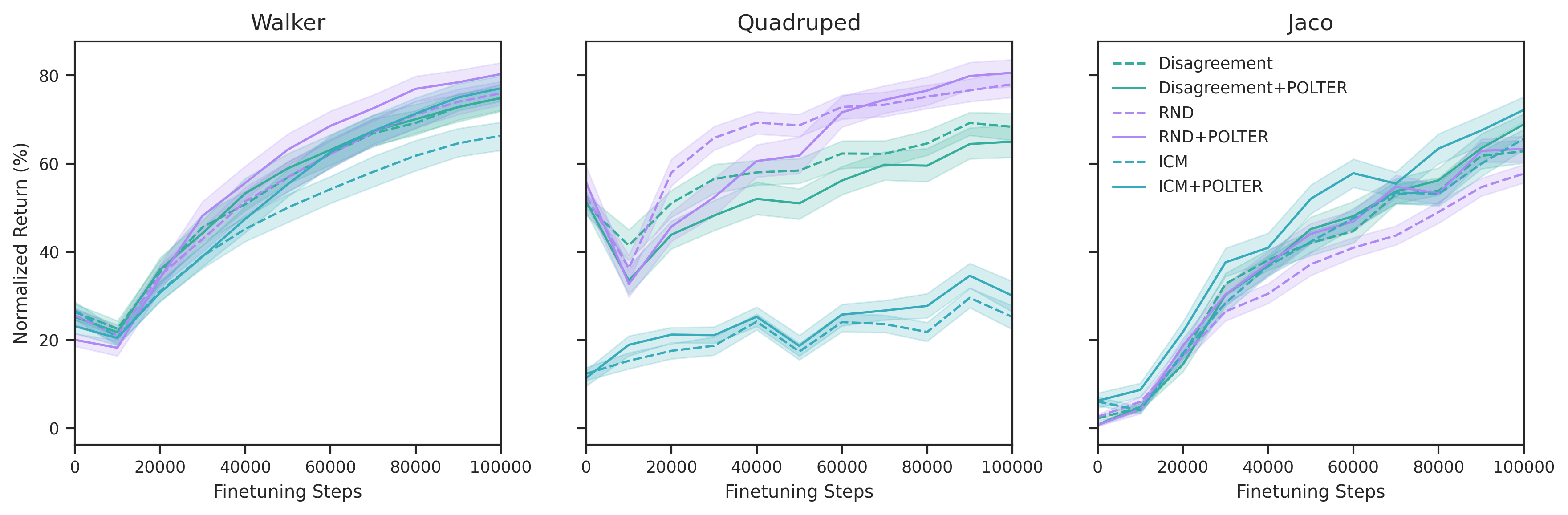}
    \includegraphics[width=0.8\textwidth]{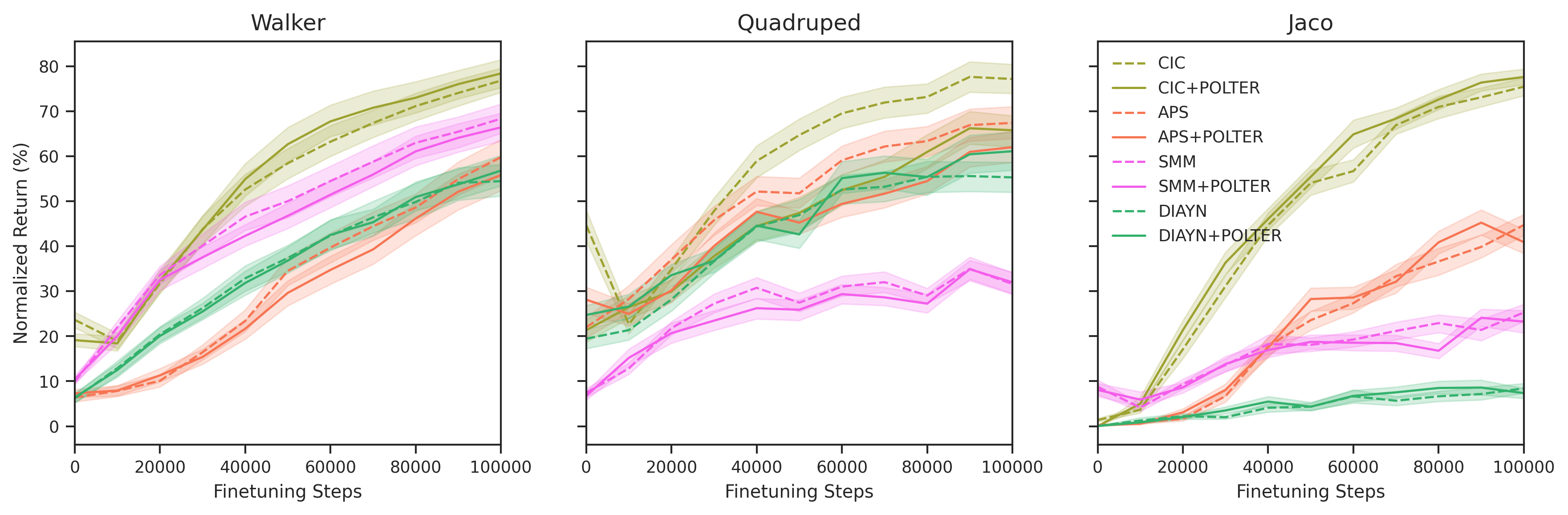}
    \caption{Finetuning curves of data-, knowledge- and competence-based algorithms after pretraining for $\SI{2}{\mega\nothing}$ steps. The shaded area indicates the standard error.}
    \label{fig:finetuning_summary}
\end{figure*}

This section provides additional results for our experiments on \acl{URLB}.
In the supplementary, we provide the raw scores.
The statistics comparing \ac{URL} algorithms with and without \polter aggregated for finetuning on \num{12} tasks across \num{10} seeds can be found in \cref{tab:agg_statistics}.
In addition we show aggregate statistics of the absolute improvement in expert performance in \cref{fig:relative_improvement_iqm_categories_perpoint} and the performance profiles~\citep{agarwalDeepReinforcementLearning2021} per \ac{URL} algorithm category in \cref{fig:perf_profiles}.
As before, we see a large improvement for data- and knowledge-based algorithms and a small or negative for competence-based algorithms.
The improvement sometimes varies strongly across seeds and tasks.
Also, we show the normalized return after finetuning from different pretraining snapshots for each domain and \ac{URL} category in \cref{fig:pretraining_summary}.
In the Jaco domain, \ac{URL} algorithms with and without \polter mostly deteriorate with an increasing number of pretraining steps.
Each category shows a different trend in each domain.
Interestingly, the competence-based algorithms \ac{SMM} and \ac{DIAYN} fail during pretraining in the Jaco domain.
In \cref{fig:finetuning_summary} we see the normalized return over finetuning steps.
\polter is mostly on par or speeds up compared to the \ac{URL} algorithm without \polter.
In total, most algorithms do not converge yet after $\SI{100}{\kilo\nothing}$ steps.

\begin{table*}[ht!]
\centering
\caption{Raw aggregate statistics following \citet{agarwalDeepReinforcementLearning2021} of evaluated \ac{URL} algorithms with and without \polter regularization. The results marked with POLTER* were obtained by tuning the regularization strength to the task domain of locomotion (Walker and Quadruped) and manipulation (Jaco).}
\label{tab:agg_statistics}
\begin{tabularx}{1\textwidth} { 
>{\raggedright\arraybackslash}l
   >{\raggedleft\arraybackslash}X
   >{\raggedleft\arraybackslash}X
   >{\raggedleft\arraybackslash}X
   >{\raggedleft\arraybackslash}X
    >{\centering\arraybackslash}X 
  }
\toprule
{} &  \ac{IQM} $\uparrow$ &  Mean $\uparrow$ &  Median $\uparrow$ &  Optimality Gap $\downarrow$ & \polter \ac{IQM} Improvement \\
Algorithm                &      &       &         &         &        \\
\midrule
\ac{ProtoRL}              & 0.56 &  0.55 &    0.52 &            0.45 & \\
\ac{ProtoRL}+\polter      & 0.77 &  0.71 &    0.65 &            0.29 & \textcolor{green}{+40\%} \\
\ac{ProtoRL}+POLTER*      & 0.79 &  0.76 &    0.80 &            0.24 & \textcolor{green}{+41\%} \\
\ac{APT}                  & 0.59 &  0.61 &    0.56 &            0.39 & \\
\ac{APT}+\polter          & 0.69 &  0.68 &    0.66 &            0.32 & \textcolor{green}{+17\%} \\
\midrule
\ac{RND}                  & 0.70 &  0.71 &    0.67 &            0.30 & \\
\ac{RND}+\polter          & 0.77 &  0.75 &    0.74 &            0.26 & \textcolor{green}{+10\%} \\
\ac{RND}+POLTER*          & 0.77 &  0.76 &    0.71 &            0.24 & \textcolor{green}{+10\%} \\
\ac{ICM}                  & 0.54 &  0.52 &    0.59 &            0.48 & \\
\ac{ICM}+\polter          & 0.63 &  0.60 &    0.65 &            0.40 & \textcolor{green}{+17\%} \\
\ac{Disagreement}         & 0.68 &  0.69 &    0.66 &            0.31 & \\
\ac{Disagreement}+\polter & 0.69 &  0.70 &    0.69 &            0.31 & \textcolor{green}{+1\%} \\
\midrule
\ac{CIC}                  & 0.78 &  0.76 &    0.74 &            0.24 & \\
\ac{CIC}+\polter          & 0.76 &  0.74 &    0.77 &            0.26 & \textcolor{red}{-2\%} \\
\ac{CIC}+POLTER*          & 0.84 &  0.81 &    0.86 &            0.20 & \textcolor{green}{+7\%} \\
\ac{DIAYN}                & 0.36 &  0.39 &    0.42 &            0.61 & \\
\ac{DIAYN}+\polter        & 0.39 &  0.42 &    0.42 &            0.58 & \textcolor{green}{+8\%}\\
\ac{SMM}                  & 0.36 &  0.42 &    0.30 &            0.58 & \\
\ac{SMM}+\polter          & 0.36 &  0.41 &    0.30 &            0.59 & $\pm$ 0\% \\
\ac{APS}                  & 0.56 &  0.58 &    0.55 &            0.42 & \\
\ac{APS}+\polter          & 0.52 &  0.53 &    0.54 &            0.47 & \textcolor{red}{-7\%} \\
\midrule
\ac{DDPG}                 & 0.55 &  0.54 &    0.56 &            0.46 & \\
\bottomrule
\end{tabularx}
\end{table*}

\end{document}